\documentclass[journal,twoside]{IEEEtran}
\usepackage{cite}
  \usepackage[pdftex]{graphicx}
  \DeclareGraphicsExtensions{.pdf,.jpeg,.png}
\usepackage[cmex10]{amsmath}
\usepackage{amsfonts}

\ifCLASSOPTIONcompsoc 
\usepackage[caption=false,font=normalsize,labelfont=sf,textfont=sf]{subfig}
\else
\usepackage[caption=false,font=footnotesize]{subfig}
\fi
\usepackage{url}
\begin{document}
\title{Spatial-Aware Dictionary Learning for Hyperspectral Image Classification}

\author{Ali~Soltani-Farani,
       Hamid R. Rabiee,~\IEEEmembership{Senior Member,~IEEE}, Seyyed Abbas Hosseini%, %Atefeh Jalali% <-this % stops a space
\thanks{Manuscript received March 18, 2013; revised July X, 2013. This work was supported in part by the AICT Research Center.}% <-this % stops a space
\thanks{A. Soltani-Farani, H. R. Rabiee, and S. A. Hosseini are with the Department of Computer Engineering, Sharif University of Technology, Tehran, Iran.}}


\markboth{IEEE TRANSACTIONS ON GEOSCIENCE AND REMOTE SENSING,~Vol.~X, No.~Y, Z~2013}%
{Soltani-Farani \MakeLowercase{\textit{et al.}}: Spatial-Aware Dictionary Learning for Hyperspectral Image Classification}



\maketitle

\begin{abstract}
%\boldmath
This paper presents a structured dictionary-based model for hyperspectral data that incorporates both spectral and contextual characteristics of a spectral sample, with the goal of hyperspectral image classification. The idea is to partition the pixels of a hyperspectral image into a number of spatial neighborhoods called contextual groups and to model each pixel with a linear combination of a few dictionary elements learned from the data. Since pixels inside a contextual group are often made up of the same materials, their linear combinations are constrained to use common elements from the dictionary.
%The basic idea is to represent a hyperspectral pixel by a linear combination of a few dictionary elements learned from the data. To incorporate contextual information, the pixels are partitioned into a number of spatial neighborhoods called contextual groups. Since spectral samples that belong to the same group are often made up of the same materials, they are constrained to belong to the same subspace. 
To this end, dictionary learning is carried out with a joint sparse regularizer to induce a common sparsity pattern in the sparse coefficients of each contextual group. The sparse coefficients are then used for classification using a linear SVM.
%by generalizing the Generalized Pareto Distribution (GPD) for multivariate data as the prior distribution of their representation coefficients. Applying MAP inference results in a joint sparse model for each contextual group.
Experimental results on a number of real hyperspectral images confirm the effectiveness of the proposed representation for hyperspectral image classification. Moreover, experiments with simulated multispectral data show that the proposed model is capable of finding representations that may effectively be used for classification of multispectral-resolution samples.
% MAP inference is used to solve for the dictionary and sparse coefficients, which leads to a two-step optimization procedure that iterates between updating the dictionary and the sparse coefficients.

\end{abstract}


% \fi
\IEEEpeerreviewmaketitle

\section{Introduction}\label{Intro}
\IEEEPARstart{N}{atural signals} are primarily modeled as members of a vector space. The dimensionality of this space is usually much higher than the number of underlying causes. This is mainly due to the inherent limitations of natural and artificial sensors, which often neglect the underlying causes of real world phenomena and therefore sample data at rates far exceeding the effective dimension of the signals. Learning these causes and thus representing a signal in a low-dimensional model is the goal of a recent trend of research known as dictionary learning \cite{Tosic11,Baraniuk10,Olshausen97}. The idea is to represent a signal by a linear combination of a few elements from a dictionary that is learned from the data. Each data point is thus represented through a sparse vector of coefficients, as a member of a low-dimensional subspace spanned by a few dictionary elements. When the dictionary is fixed this process is commonly known as sparse coding. Dictionary learning has achieved great success in signal reconstruction tasks such as compression \cite{Aharon06} and denoising \cite{Elad06}. More recently, it has also been applied to discriminative tasks such as classification \cite{Yang09,Zhou10,Culpepper11,Zhang10,Mairal12} and clustering \cite{Ramirez10,Elhamifar11} with state-of-the-art results.

A hyperspectral image is a collection of pixels that represent a given scene or object, where pixels represent the reflected solar radiation from the Earth's surface in many narrow spectral bands \cite{Plaza09}. At each pixel, the spectral features form a vector whose elements correspond to the narrow bands covering visible to infrared regions of the spectrum. The wealth of data provided by hyperspectral imagery (HSI) has promoted its application in many domains such as agriculture \cite{Bannari06,Larsolle07}, defense \cite{Banerjee06,Eismann09}, and environmental management  \cite{Lawrence06,Zomer09}.
The reflectance spectra of a pixel are influenced by a number of factors. Apart from measurement noise caused by variation in illumination and viewing angle, and environmental effects such as aerosols and moisture, the spectral features of a pixel are determined by the material present at the given pixel and its surrounding area. Due to the spatial resolution of the imaging device, scattering from the local scene, and material mixtures within a pixel, each pixel is often composed of a number of different materials plus noise \cite{Gomez07,Bioucas-Dias12}. Spectral unmixing is the process of identifying the pure materials present in the mixture, called \emph{endmembers} and their respective \emph{abundances}. The linear mixture model (LMM) which is commonly used for unmixing assumes that each pixel, $x$ is composed of a linear combination of endmembers, $D=[d_1,\ldots,d_K]$ plus an additive noise, $\epsilon$, i.e. 
\begin{equation}
x=Dy+\epsilon
\end{equation}
where the fractional abundances, $y$ are commonly assumed to be nonnegative and sum to unity. This is essentially the idea encouraged by dictionary learning with one difference, that in dictionary learning the fractional abundances are mainly assumed to be sparse. The aim of dictionary learning is to reduce the error in representing each signal while inducing sparsity in the representation coefficients. This is commonly accomplished through a formulation such as:
\begin{equation}\label{DL_GeneralForm}
\arg\min_{D,y_i\in\mathcal{C}}\sum_{i=1}^N\left(\frac{1}{2}\|x_i-Dy_i\|_2^2 +\gamma\mathcal{S}(y_i)\right)
\end{equation}
where $N$ is the number of signals available for training, $\mathcal{S}(y)$ is an sparsity inducing regularizer such as the well known $\ell_1$ norm, $\gamma$ is a regularization parameter balancing representation error with representation complexity, and $\mathcal{C}$ represents constraints on the sparse coefficients.

Recently, inspired by the ability of dictionary learning to model high-dimensional data and its potential to learn high-level information from the training samples \cite{Tosic11,Olshausen97}, sparse coding and dictionary learning have been used for spectral unmixing with encouraging results. The sparse unmixing approach proposed in \cite{Iordache11} assumes that a set of pure spectral signatures are available which compose the dictionary. The fractional abundances are estimated using sparse coding with an $\ell_1$ regularizer, taking into account the fact that a few number of endmembers contribute to a given pixel. This approach was later extended in \cite{Iordache12_1}, where it is assumed that all pixels in an image have fractional abundances with a common sparsity pattern. This is achieved by using a joint sparse regularizer to also decrease the total number of endmembers activated for an image. Since spectral libraries are often composed of groups of spectral signatures, \cite{Iordache11_2} proposes a group Lasso formulation to exploit this fact. Motivated by the observation that pixels in a hyperspectral image are usually surrounded by similar pixels, \cite{Iordache12_2} include the total variation (TV) regularization in the sparse unmixing formulation to encourage smooth variation in the fractional abundance of each endmember among adjacent pixels. The aforementioned approaches assume a library of pure spectra is given a priori which make up the dictionary (i.e. $D$ is fixed in (\ref{DL_GeneralForm})). Selecting endmembers from the data has also been attempted both manually, based on the similarity between eigenvectors of the scene and the data \cite{Bateson96} or automatically, based on a measure of modeling quality \cite{Gomez07}. In contrast, \cite{Charles11, Castrodad11,Greer12} attempt to learn the set of spectral endmembers using dictionary learning. In \cite{Charles11} the dictionary is learned from unsupervised training data by considering a probabilistic LMM framework, wherein the additive noise is assumed to be Gaussian and the fractional abundances are i.i.d. Laplacian and constrained to be nonnegative. Experimental results show that the learned dictionary elements are similar to the material spectra available in the scene and may be used to infer samples with HSI-resolution from multispectral-level measurements. The work of Greer \cite{Greer12} differs from \cite{Charles11} in that the dictionary is assumed to have full rank and the fractional abundances must sum to unity. Hence, the $\ell_1$ sparsity regularizer used in previous work no longer applies. A recent survey of different approaches for hyperspectral unmixing is presented in \cite{Bioucas-Dias12}.
%Unlike the above work, in \cite{Xing12} dictionary learning is used to remove noise or infer missing data from HSI, by considering both spectral and spatial features.

The high spatial and spectral resolution of a hyperspectral image provides the potential for each pixel to be accurately and robustly labeled as one of a known set of classes. Hyperspectral image classification has been applied to both urban \cite{Benediktsson05} and  agricultural \cite{Camps03} scenery. Various methods have been developed for this application. Among them are supervised techniques such as maximum likelihood and Bayesian classifiers \cite{Landgrebe03}, decision trees \cite{Goel03}, neural networks \cite{Yang99}, and support vector machines (SVM) \cite{Gualtieri99,Melgani04}.  Semisupervised learning based on graph construction \cite{Camps07} and transductive SVMs \cite{Bruzzone06} have also been proposed which take advantage of both labeled and unlabeled samples for classification. Inspired by the fact that pixels in a hyperspectral image are often surrounded by pixels of the same class, recent methods have focused on both spectral and contextual characteristics of HSI. The composite-kernel SVM \cite{Camps06} makes use of Mercer's theorem to construct kernels composed from a spectral kernel and a contextual kernel. In particular, the weighted sum of the spectral and contextual kernels has been successful in classifying images with limited training samples. The graph kernel SVM \cite{Camps10} incorporates both spectral and contextual characteristics simultaneously into a recursive graph kernel,
%The kernel at depth 1 is defined for two given pixels by considering a neighborhood around each pixel and averaging the similarities between all pairs of pixels inside those neighborhoods. At consecutive depths the similarity between a pair of pixels is determined by the previously computed kernel. 
and is also effective for small training sample sizes.  Due to its recent success in discriminative tasks with small training data \cite{Yang09,Zhou10,Culpepper11,Zhang10,Mairal12,Ramirez10,Elhamifar11}, dictionary learning and sparse coding have also been applied to hyperspectral image classification. A sparsity-based model is proposed in \cite{Chen11} where a test spectral sample joined with its surrounding pixels is represented by a few training samples from a fixed training dictionary. The test pixel is then labeled as the class whose training samples have the largest contribution in representing the pixel and its surrounding neighbors. Although the main focus of \cite{Charles11} is spectral unmixing, the authors demonstrate that using the fractional abundances instead of the raw spectral features improves the classification accuracy of the linear SVM for small training sets. In the approach proposed by Castrodad et al. \cite{Castrodad11}, a dictionary is learned for each class of hyperspectral data in a supervised manner. For classification, the different dictionaries are concatenated to form a single dictionary. The sparse code (fractional abundance) corresponding to a new pixel is calculated using the sparse unmixing formulation accompanied by a spectral-spatial regularizer to enforce smooth variations in the sparse codes for neighboring pixels. The new pixel is then labeled as the class whose dictionary produces the lowest representation error and complexity.

The work discussed above for hyperspectral unmixing and classification enjoy a number of common and individual advantages and pose a number of remaining challenges which motivate our work. Specifically, we focus on a simple yet efficient approach to  learn a dictionary for hyperspectral data that can incorporate contextual information, with the aim of hyperspectral image classification. Of the approaches for spectral unmixing, some assume that a set of pure spectral signatures are available which may compose the dictionary \cite{Iordache11,Iordache11_2,Iordache12_1,Iordache12_2}. Chen et al. \cite{Chen11} use the complete set of training data as the dictionary, but with the aim of classification. Similar to \cite{Charles11, Castrodad11, Greer12} we learn a dictionary using the training data, which is less complex (i.e. consists of fewer atoms) yet is more effective for classification (Section \ref{Experiments}). In terms of incorporating contextual information, as we discuss in Section \ref{SpectralORContextual},\cite{Camps06, Camps10, Chen11} employ a window centered at the pixel of interest to gather contextual information. This hinders the potential for parallel computation \cite{Plaza09} and the methods tend to use only contextual information. In fact, as we shall see in the experimental results of Section \ref{Experiments}, SVM classification using the weighted sum of the spectral and contextual kernels \cite{Camps06} achieves highest accuracy when the spectral kernel is given zero weight. The methods of \cite{Chen11, Camps10} are also based on the contextual characteristics of HSI since the pixel and its surrounding neighbors are indistinguishable to the classifying process. Of the dictionary-based approaches,  Iordache et al. \cite{Iordache12_2} and Castrodad et al. \cite{Castrodad11} incorporate contextual information by augmenting (\ref{DL_GeneralForm}) with a regularization term that enforces smooth variation in the sparse representation for neighboring pixels. In \cite{Iordache12_2} the dictionary is fixed and \cite{Castrodad11} learns the dictionary without the proposed regularization term. This is perhaps in regard of the complex optimization procedure, which is in turn due to the fact that the sparse representations of different pixels can not be computed independent of each other. In contrast, we attempt to learn the dictionary and incorporate contextual information simultaneously, yet our optimization is simple and amenable to parallel computations. 

In this paper, we propose a structured dictionary-based model for hyperspectral data that is impervious to the aforementioned issues and incorporates both spectral and contextual characteristics of a spectral sample into the sparse set of coefficients. The idea is to partition the pixels of a hyperspectral image into a number of spatial neighborhoods called contextual groups and to model each pixel with a linear combination of a few elements from a dictionary. Since pixels inside a contextual group are often made up of the same materials, their linear combinations are constrained to use common elements from the dictionary. Equivalently, they belong to the same subspace and their sparse coefficients have a common sparsity pattern. This is realized by using a joint sparsity inducing regularizer in the dictionary learning formulation of (\ref{DL_GeneralForm}). We also show how this model may be viewed from a probabilistic perspective by building upon the basic probabilistic framework introduced in \cite{Olshausen97} and employed in \cite{Charles11}. Solving for the dictionary and sparse coefficients leads to a two-step optimization procedure that iterates between updating the dictionary and the sparse coefficients. Each step is a convex optimization which is well known in the literature and for which efficient solutions exist. Recent work in the field of computer vision using sparse coding and dictionary learning techniques \cite{Yang09, Zhou10,Culpepper11} has shown that the extracted features therein are discriminative enough to be well classified using a simple classifier such as linear SVM. Motivated by these recent findings, we employ a linear SVM to classify the sparse representation corresponding to each pixel. Extensive experiments on real hyperspectral images are provided to assess the properties of the proposed model.

To summarize, we make the following main contributions: (i) We show that the proposed model is capable of incorporating both spectral and contextual characteristics of a spectral sample into the sparse set of coefficients. Extensive experiments on three hyperspectral datasets show that the inferred sparse coefficients are discriminative enough to be classified with state-of-the-art accuracy using a linear SVM. (ii) Charles et al. \cite{Charles11} show that the sparse coding model accompanied by an HSI dictionary can be used to infer HSI resolution data from simulated multispectral imagery (MSI). We classify the sparse representations retrieved from simulated MSI-resolution data and show that our model is capable of finding representations that may effectively be used for classification of MSI-level samples. 
%We compare the fractional abundances retrieved from simulated MSI-level resolution data to those inferred from HSI resolution samples and show that our model is capable of finding similar fractional abundances that may effectively be used for classification of MSI-level samples.
(iii) Moreover, compared to dictionary-based hyperspectral image classification methods \cite{Chen11, Castrodad11}, we use a smaller number of dictionary elements for classification and show that our method is amenable to efficient parallel processing.

The remainder of this paper is organized as follows. Section II provides the necessary background on dictionary learning and introduces the structured dictionary-based model for hyperspectral data. To gain further insight, the models are also analyzed from a probabilistic point of view. The details of both basic and structured models and their application to HSI classification is discussed in Section III. To demonstrate the effectiveness of the proposed model, extensive experimental results on several hyperspectral images are reported and analyzed in Section IV. Section V concludes this paper and discusses paths for future research.

\section{Dictionary-Based Models for HSI}\label{Section_II}
In this section, we first provide a brief background on the general dictionary learning paradigm followed by a short analysis of the dictionary learning setting employed in \cite{Charles11,Castrodad11}. We then customize the general model into a structured joint-sparse model tailored for hyperspectral data. We also describe how learning the parameters of these models leads to convex programs for updating the dictionary and sparse representations. In the sequel, lower case and capital letters are used for vectors ($x$) and matrices ($X$) respectively. Random variables are written in boldface letters.

%In this section, we first review the general dictionary learning paradigm from a probabilistic view point and derive the simple dictionary learning model \cite{Olshausen97} by assuming that the sparse representations are independent realizations of the i.i.d. Generalized Pareto Distribution (GPD). The general model is then customized into a structured joint-sparse model tailored for hyperspectral data, by introducing a multivariate GPD. We also show that inference on these models leads to convex programs for updating the dictionary and sparse representations. In the sequel, lower case and capital letters are used for vectors ($x$) and matrices ($X$) respectively. Random variables are written in boldface letters.
\subsection{Dictionary learning: General Paradigm}
Let $\mathcal{X}\subset\mathbb{R}^M$ denote the set of signals of interest, e.g. the pixels of a hyperspectral image. Given ${x_1,\ldots,x_N}\in\mathcal{X}$, the fundamental goal of dictionary learning is to find a set of atomic signals $D=[d_1,\ldots,d_K]$ that form the building blocks of $\mathcal{X}$, in the sense that any $x\in\mathcal{X}$ is represented by a linear combination of a few of these atoms, i.e.
\begin{equation}\label{LMM_DL}
x=Dy+\epsilon
\end{equation}
where $\epsilon$ is a small residual due to modeling $x$ in a linear manner with the sparse representation vector $y\in\mathbb{R}^K$. Depending on the particular application, the desired accuracy and complexity, and the nature of the signals, dictionary learning may take different forms, yet is often a regularized least squares optimization:
\begin{equation}\label{DL_General}
\arg\min_{D,Y}\frac{1}{2}\|X-DY\|_F^2 +\gamma\mathcal{S}(Y)\end{equation}
where $X=[x_1,\ldots,x_N]$ , $Y=[y_1,\ldots,y_N]$, and $\|.\|_F$ denotes the Frobenius norm of a matrix. The regularizer, $\mathcal{S}(Y)$ is mainly sparsity inducing but may also induce other forms of a priori knowledge and $\gamma$ is the regularization parameter. The formulation of (\ref{DL_General}) may also be viewed from a probabilistic perspective. Assuming that the residual vectors are independent zero-mean Gaussians with covariance matrix $\sigma^2I$ and using the Bayes rule, the posterior is given by:
\begin{equation}
P(D,Y|X)\propto P(X|D,Y)P(D)P(Y)
\end{equation}
where $P(X|D,Y)=\prod_{i=1}^N \mathcal{N}(x_i|Dy_i,\sigma^2I)$.  For convenience, it is usually assumed that $P(D)$ is uniform, leaving $P(Y)$ as the only means of conveying one's knowledge about $X$. If $D,Y$ are estimated from the \emph{maximum a posteriori} or \emph{MAP estimate} we arrive at a form similar to (\ref{DL_General}):
\begin{equation}\label{MAP_General}
\arg\min_{D,Y}\frac{1}{2}\|X-DY\|_F^2 -\sigma^2\log P(Y)
\end{equation}

%Under the Bayesian framework, one may convey uncertainty about $X=[x_1,\ldots,x_N]$ through the dictionary $D$, the sparse representation matrix $Y=[y_1,\ldots,y_N]$, and the residual matrix $E=[\epsilon_1,\ldots,\epsilon_N]$, distributed according to $P(D)$, $P(Y)$, and $P(E)$ respectively. Assuming that the residual vectors are independent zero-mean Gaussians with covariance matrix $\sigma^2I$ and using the Bayes rule, the posterior is easily computed as:
%\begin{equation}
%P(D,Y|X)\propto P(X|D,Y)P(D)P(Y)
%\end{equation}
%where $P(X|D,Y)=\prod_{i=1}^N \mathcal{N}(x_i|Dy_i,\sigma^2I)$.  $D,Y$ can be estimated from the \emph{maximum a posteriori} or \emph{MAP estimate} given by:
%\begin{equation}\label{MAP_General}
%\arg\min_{D,Y}\frac{1}{2\sigma^2}\|X-DY\|_F^2 -\log P(D) -\log P(Y)
%\end{equation}
%where $\|.\|_F$ denotes the Frobenius norm of a matrix. For convenience it is usually assumed that $P(D)$ is uniform, leaving $P(Y)$ as the only means of conveying one's knowledge about $X$.

\subsection{Dictionary Learning: Basic Setting}\label{DL_SimpleSetting}
In its simplest form, dictionary learning is performed with an sparsity inducing regularizer that acts independently on $y_1,\ldots,y_N$ and also on the elements of these vectors, $y^{(1)},\ldots,y^{(K)}$. From a probabilistic perspective it is assumed that the sparse representations, $y_1,\ldots,y_N$, are independent realizations of a random vector $\bf{y}$ and that the elements of this random vector, $\bf{y}^{(1)},\ldots,\bf{y}^{(K)}$, are also independent random variables distributed according to a common pdf, $P(y)$. Traditionally, a Laplacian distribution has been attractive as it leads to the well known Lasso or $\ell_1$ minimization\cite{Tibshirani96} for recovering $Y$:
\begin{equation}\label{DL_Simple}
\arg\min_{D,Y}\frac{1}{2}\|X-DY\|_F^2 +\gamma\sum_{i=1}^N \|y_i\|_1
\end{equation}
This is the form employed by \cite{Charles11,Castrodad11} for dictionary learning with the added constraint that all elements of $D$ and $Y$ are nonnegative. To avoid the trivial solution in which the rows of $Y$ tend to zero while the dictionary atoms become prohibitively large, the above optimization is solved with the constraint that $\|d_i\|\le1$ for $i=1,\ldots,K$. This could have been  accounted for in the prior, $P(D)$, as in \cite{Kreutz03}, but would lead to slower training algorithms. Before discussing how the above optimization is solved, we briefly note that it was recently shown in \cite{Baraniuk10} that realizations of the i.i.d. Laplacian are not sparse\footnote{To be accurate, realizations of the i.i.d. Laplacian do not exhibit a power law decay or equivalently are not compressible. See \cite{Baraniuk10} for details.}. This raises the question as to why solutions of $\ell_1$ minimization are sparse and what prior distribution may be used to express sparsity. The authors show that the i.i.d. zero-mean Generalized Pareto Distribution (GPD):
\begin{equation}\label{GPD}
P_{\text{GPD}}(y)=\frac{\lambda}{2}\left(1+\frac{\lambda|y|}{q}\right)^{-(q+1)}
\end{equation}
with order q and shape parameter $\frac{1}{\lambda}$ is a valid sparsity inducing prior for $\bf{y}$, and the Lasso minimizes an upper bound on its MAP estimation. We are interested in this distribution because we later wish to generalize it to model the dependency between contextually related pixels. This will help us set the regularization parameter, $\gamma$ in (\ref{DL_GeneralForm}), so that contextual groups of different size are modeled with similar complexity (see Section \ref{DL_HSI} for more details). 
The above optimization is convex in either $D$ or $Y$ but not in both. A common two-step strategy is used for this problem. These steps are iterated until convergence. 

\subsubsection{Sparse Coding}
In this step, $D$ is fixed and the optimization is solved with regard to $Y$. The objective function in (\ref{DL_Simple}) is separable and may be solved for each $y_i$ independently by:
\begin{equation}\label{L1}
\arg\min_{y_i} \frac{1}{2}\|x_i-Dy_i\|_2^2 +\gamma \|y_i\|_1
\end{equation}
which is known as the Lasso, BPDN, or $\ell_1$ minimization. Several efficient algorithms have been proposed to solve (\ref{L1}) \cite{Osborne00,Efron04,Lee07}, among which we use the implementation of \cite{Efron04} provided by the SPAMS toolbox \cite{Mairal10,SPAMS12}.
\subsubsection{Dictionary Update}\label{BCD_dict_update}
For the dictionary update step, $Y$ is fixed and the optimization becomes
\begin{eqnarray}
\arg\min_{D}& \frac{1}{2} \|X-DY\|_F^2  \\ \nonumber
\text{s.t. } \forall i&\|d_i\|_2\le1
\end{eqnarray}
which is quadratic in $D$. The gradient of the objective function equals $DYY^T-XY^T$ which is zero for $D=XY^T(YY^T)^{-1}$. To account for the constraints it suffices to project atoms with larger than unit norm onto the unit $\ell_2$ ball. This is the solution provided by \cite{Engan99}, a straight forward approach which may suffer from  calculating the inverse of $YY^T$. The method proposed in \cite{Lee07} solves the dual problem in an iterative manner and \cite{Olshausen97} employs a steepest descent strategy. An online dictionary learning algorithm is proposed in \cite{Mairal10} which is suitable for problems with many training data. We use a Block Coordinate Descent (BCD) strategy that updates the dictionary atoms iteratively. Since the objective function is strongly convex, BCD is guaranteed to achieve the unique solution. The objective function for the $j$'th atom may be written as
%\begin{equation}
%\frac{1}{2} \left\|X-\sum_{i=1}^Kd_i{Y_T^i}^T\right\|_F^2=
$\frac{1}{2} \left\|R_j-d_j{Y_T^j}^T\right\|_F^2$,
%\end{equation}
where $Y_T^j$ is the $j$'th row of $Y$ in column form and $R_j=X-\sum_{i\ne j}d_i{Y_T^i}^T$. Keeping only the terms in $d_j$, a little algebra yields:
\begin{eqnarray}
\arg\min_{d_j}& \frac{1}{2} \left\|d_j-\frac{R_jY_T^j}{\|Y_T^j\|_2^2}\right\|_2^2  \\ \nonumber
\text{s.t. } &\|d_j\|_2\le1
\end{eqnarray}
the solution of which is
\begin{equation}
d_j=\text{proj}_{\ell_2}\left(\frac{R_jY_T^j}{\|Y_T^j\|_2^2}\right)
\end{equation}
where $\text{proj}_{\ell_2}(x)$ denotes the projection of $x$ in the unit $\ell_2$ ball.

\subsection{Dictionary Learning for HSI}\label{DL_HSI}
In the dictionary learning formulation of (\ref{DL_Simple}), the sparse representations are assumed to be independent. This simplifies the sparse coding step of dictionary learning since the objective function becomes separable in $y_1,\ldots,y_N$. In the case of HSI, this simple setting ignores the large spatial correlation of HSI pixels. To overcome this problem, we partition the pixels into a number of spatial neighborhoods called contextual groups. Pixels that belong to the same contextual group are often made up of the same material; accordingly we assume that their representations use a common set of atoms from the dictionary. Thus, the sparse representations of pixels that belong to the same group are no longer independent. In Section \ref{Define_Contextual} we discuss how the contextual groups are defined. Let $\{\mathcal{G}_1,\ldots,\mathcal{G}_g\}$, denote the set of contextual groups, defined as a partition on $\{x_1,\ldots,x_N\}$, and $X_{\mathcal{G}_i}=[x_{i,1},\ldots,x_{i,|\mathcal{G}_i|}]$ represent the members of $\mathcal{G}_i$ written as columns of the matrix $X_{\mathcal{G}_i}$. Accounting for the above assumption, the model in (\ref{LMM_DL}) may now be written as
\begin{equation}\label{Joint_LMM_DL}
X_{\mathcal{G}_i}=DY_{\mathcal{G}_i}+E_{\mathcal{G}_i}
\end{equation}
where the columns of $Y_{\mathcal{G}_i}$ and $E_{\mathcal{G}_i}$ are respectively the sparse representations and error vectors corresponding to the spectral samples in the columns of $X_{\mathcal{G}_i}$. $Y_{\mathcal{G}_i}$ is a row sparse matrix i.e. its columns have a common sparsity pattern. To learn the dictionary and sparse representations, we employ the $\ell_2$/$\ell_1$ convex joint sparsity inducing regularizer in (\ref{DL_General}) to arrive at:
\begin{eqnarray}\label{DL_L2GPD}
\arg\min_{D,Y}&\frac{1}{2} \|X-DY\|_F^2 + \sum_{i=1}^g \gamma_{_{\mathcal{G}_i}}\left\|Y_{\mathcal{G}_i}\right\|_{2,1}\\\nonumber
\text{s.t. } \forall i &\|d_i\|_2\le1
\end{eqnarray}
where $\gamma_{_{\mathcal{G}_i}}$ is the regularization parameter for the $i$th group and $\|Y\|_{2,1}$ is the $\ell_2$/$\ell_1$ norm defined as the sum of the $\ell_2$ norms of the rows of $Y$.

From a probabilistic viewpoint, this is equivalent to assuming that $Y_{\mathcal{G}_1},\ldots,Y_{\mathcal{G}_g}$ are independent realizations of a random matrix, $\bf{Y}_{\mathcal{G}}$, the rows of which are independent random vectors distributed according to a common pdf, $P(Y_{\mathcal{G},T})$.  In other words, we consider the dependency between members of a contextual group, but similar to the setting in (\ref{DL_Simple}), we don't model the dependency between the dictionary atoms. To define $P(Y_{\mathcal{G},T})$, we propose to generalize the GPD for vector data as:
\begin{equation}\label{L2GPD}
P(Y_{\mathcal{G},T})=C_{\mathcal{G}}\left(1+\frac{\lambda_{\mathcal{G}}\|Y_{\mathcal{G},T}\|_2}{q_{_\mathcal{G}}}\right)^{-(q_{_{\mathcal{G}}}+1)}
\end{equation}
where $C_\mathcal{G}$
%\begin{equation}
%C_\mathcal{G}=\frac{\Gamma(\frac{|\mathcal{G}|}{2})\lambda_{\mathcal{G}}^{|\mathcal{G}|}\prod_{i=1}^{|\mathcal{G}|-1}(q_{_{\mathcal{G}}}-i)}{2\pi^{|\mathcal{G}|}q_{_{\mathcal{G}}}^{|\mathcal{G}|-1}(|\mathcal{G}|-1)!}
%\end{equation}
is the normalization factor, and $q_{_\mathcal{G}}>|\mathcal{G}|-1$. We call this distribution $\ell_2$-GPD, since the $\ell_2$ norm of the row $Y_{\mathcal{G},T}$, is distributed according to (\ref{GPD}). This results in row-sparse matrices $Y_{\mathcal{G}}$, which is equivalent to the above joint sparse model for each contextual group. Applying MAP estimation as before shows that (\ref{DL_L2GPD}) minimizes an upper bound on the MAP objective function with $\gamma_{_{\mathcal{G}_i}}=\sigma^2\lambda_{\mathcal{G}_i}(\frac{1}{q_{_{\mathcal{G}_i}}}+1)$. we set $q_{_\mathcal{G}}=|\mathcal{G}|$ and use $\lambda_\mathcal{G}=\sqrt{|\mathcal{G}|}$ to achieve similar sparsity in groups of different size. An interesting study, beyond the scope of our work, would be to include the representation complexity of each group in the model by inferring this parameter from the data.
We should note that MAP estimation with a Laplacian prior on the $\ell_2$ norm of each row of $Y_\mathcal{G}$ or an equivalent hierarchical prior as in \cite{Martins11} would also lead to (\ref{DL_L2GPD}) but suffers from the same issue discussed in Section \ref{DL_SimpleSetting}, i.e. its realizations are not row sparse matrices.

%(\ref{DL_L2GPD}) may had we considered a Laplacian prior on the $\ell_2$ norm of each row of $Y_\mathcal{G}$ or an as in \cite{Martins11}. Unfortunately, this distribution suffers from the same issue discussed in Section \ref{DL_SimpleSetting}, i.e. its realizations are not row sparse matrices. Moreover, $\gamma_{_{\mathcal{G}_i}}$  would have differed from what is derived here, which would not permit us to set this parameter in order to achieve similar sparsity in groups of different size. An interesting study, beyond the scope of our work, would be to include the representation complexity of each group in the model by inferring this parameter from the data.

Similar to (\ref{DL_Simple}), the above optimization is convex in either $D$ or $Y$, but not in both. We use the same two-step strategy of iterative sparse coding and dictionary update. Although, the dictionary update step need not be changed, the sparse coding phase can no longer be solved independently for each $y_i$. The objective function of this step is still separable and can be solved for each $Y_\mathcal{G}$ by:
\begin{equation}\label{JSCL21}
\arg\min_{Y_\mathcal{G}} \frac{1}{2} \|X_\mathcal{G}-DY_\mathcal{G}\|_F^2 + \gamma_{_{\mathcal{G}}}\left\|Y_{\mathcal{G}}\right\|_{2,1}
\end{equation}
where the columns of $X_\mathcal{G}$ are the spectral samples corresponding to the sparse representations in the columns of $Y_\mathcal{G}$. This is the well known convex formulation of the joint sparse recovery \cite{Baraniuk10} or simultaneous sparse approximation \cite{Tropp06} problem, also known as the multiple measurement vector (MMV) \cite{Cotter05} problem in the compressed sensing community. Other convex formulations for this problem follow the general $\ell_q$/$\ell_p$ form with $q\ge1$ and $p\le1$, among which $\ell_2$/$\ell_1$ and $\ell_\infty$/$\ell_1$ formulations are more widely used. We adhere to the $\ell_2$/$\ell_1$ form for two reasons. First, the objective function of (\ref{JSCL21}) is strongly convex for $|\mathcal{G}|>1$, and therefore has a unique solution. Several algorithms exist that efficiently solve this problem \cite{Cotter05,Malioutov05,Lu11,Rakotomamonjy11}. Second, we arrived at this formulation within a probabilistic framework by introducing the $\ell_2$-GPD prior of (\ref{L2GPD}). The $\ell_2$ norm in this formulation treats all entries of $Y_{\mathcal{G},T}$ equally while an $\ell_\infty$ norm is solely determined by the maximum absolute value in $Y_{\mathcal{G},T}$. In other words, the $\ell_2$ norm treats all spectral samples alike, while the $\ell_\infty$ norm may bias $Y_\mathcal{G}$ towards using the dictionary atoms for which a corresponding spectral sample with very high similarity exists.

Among the algorithms proposed to solve (\ref{JSCL21}), Malioutov, et al.\cite{Malioutov05} show that the optimization may be posed as Second Order Cone Programming (SOCP) for which off-the-shelf optimizers are available. The inner loops of SOCP are computationally expensive and the algorithm is only suitable for small-size problems \cite{Malioutov05,Lu11}. The work of Lu, et al. \cite{Lu11} extends the Alternating Directions Method (ADM) of \cite{Yang11} to solve MMV recovery. Although the proposed algorithm is quite fast within an acceptable solution accuracy, there is no guarantee that each iteration of the algorithm will reduce the objective function. In fact, as is observed in \cite{Lu11} and explained by Yang et al. \cite{Yang11}, if $X_\mathcal{G}\ne DY_\mathcal{G}$, the objective function may increase after some iterations. This is particularly important for our scenario of dictionary learning where a nonzero residual due to linear modeling is inevitable. We employ the regularized M-FOCUSS algorithm of Cotter et al. \cite{Cotter05} which is simple, efficient, and also guaranteed to reduce the objective function in each iteration. The algorithm works by estimating the $\ell_2$ norm of each row of $Y_\mathcal{G}$, and then updating $Y_\mathcal{G}$ based on that estimate. Setting the gradient of the objective function in (\ref{JSCL21}) to zero, we arrive at:
\begin{equation}
\Lambda D^TDY_\mathcal{G} - \Lambda X_\mathcal{G}^TD + \gamma_{_\mathcal{G}}Y_\mathcal{G}=0
\end{equation}
where $\Lambda = \text{diag}(\|Y_{\mathcal{G},T}^i\|_2)$. Using the Searle identity \cite{Searle82} suggests the following update equation
\begin{equation}\label{MFOCUSS_update}
Y_\mathcal{G} = \Lambda D^T (D\Lambda D^T + \gamma_{_\mathcal{G}}I)^{-1}X_\mathcal{G}
\end{equation}
where $\Lambda$ is computed using the previous estimate of $Y_\mathcal{G}$. The algorithm may be initialized from any random point for which $\forall i, \|Y_{\mathcal{G},T}^i\|_2\ne0$ and is terminated when the difference between consecutive estimates of $Y_\mathcal{G}$ is smaller than some threshold. Another solution for the optimization in (\ref{JSCL21}) is based on a BCD strategy that iteratively solves (\ref{JSCL21}) for each row of $Y_\mathcal{G}$. This solution is derived similar to the BCD-based dictionary learning procedure explained in Section \ref{BCD_dict_update}, hence for convenience we only provide the update rule for the $j'th$ row of $Y_\mathcal{G}$,
\begin{equation}\label{BCD_update}
Y_{\mathcal{G},T}^j=\left(1-\frac{\gamma_{_\mathcal{G}}}{\|R_j^Td_j\|_2}\right)_+\frac{R_j^Td_j}{\|d_j\|_2^2}
\end{equation}
in which $(x)_+ = \max(x,0)$. Although applying (\ref{BCD_update}) for each row is less expensive than applying (\ref{MFOCUSS_update}), we have empirically found regularized M-FOCUSS to need far fewer iterations for convergence than the BCD-based algorithm.

\section{Dictionary-based Classification of HSI}\label{Section_III}
In this section, we discuss how the models introduced in Section \ref{Section_II} are used for hyperspectral image classification. The basic idea is to learn a dictionary from the data and approximate each pixel with a linear combination of a few dictionary elements. The coefficients of this linear combination form a sparse vector that is used to classify the corresponding pixel. We briefly discuss how (\ref{DL_Simple}) is used in \cite{Charles11} to classify hyperspectral samples based on their spectral features. We then consider, as a basic extension, to learn the dictionary based on contextual data alone. Finally, we explore how the structured model is learned from spectral data partitioned into predefined contextual groups.

\subsection{Spectral/Contextual Dictionary Learning}\label{SpectralORContextual}
To learn the dictionary $D$ from spectral data, let $x_1,\ldots,x_N$ denote the spectral representation of the training data with respective labels $l_1,\ldots,l_N$. Applying the dictionary learning formulation of (\ref{DL_Simple}) to these samples yields corresponding sparse representations $y_1,\ldots,y_N$ and the dictionary $D$. In \cite{Charles11} it is also assumed that the dictionary elements and sparse coefficients are nonnegative. A linear SVM is trained on the sparse representations and their corresponding labels $l_1,\ldots,l_N$. Given a new spectral sample $x$, sparse coding is applied as in (\ref{L1}) to find the corresponding sparse representation $y$, which is then classified using the trained linear SVM to find the corresponding label $l$. This is a very straightforward method for applying dictionary learning to HSI classification. Like other classification methods that only make use of spectral characteristics of HSI, it has limited classification capability, which we shall observe in the experimental results of Section \ref{Experiments}.

%\begin{figure}[!t]
%\centering
%\includegraphics[width=1.5in]{Neib_S}
%% where an .eps filename suffix will be assumed under latex, 
%% and a .pdf suffix will be assumed for pdflatex; or what has been declared
%% via \DeclareGraphicsExtensions.
%\caption{Block of a hyperspectral image. Shaded areas show windows centered at pixels $s_i$ and $s_j$.  The contextual representation of a pixel is computed from pixels inside its window.}
%\label{Neib_S}
%\end{figure}

\begin{figure}[!t]
\centerline{\subfloat[]{\includegraphics[width =1.7in]{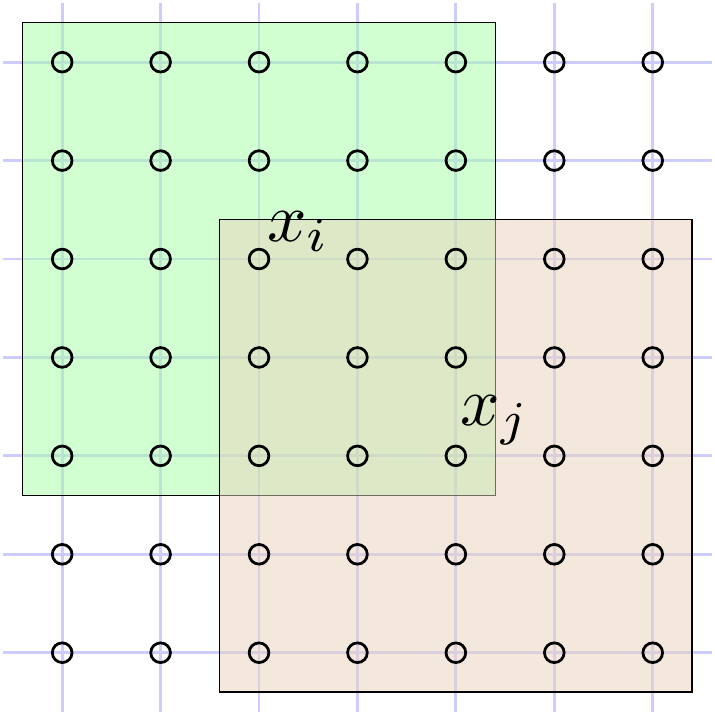}
\label{Neib_S}}
\hfil
\subfloat[]{\includegraphics[width=1.7in]{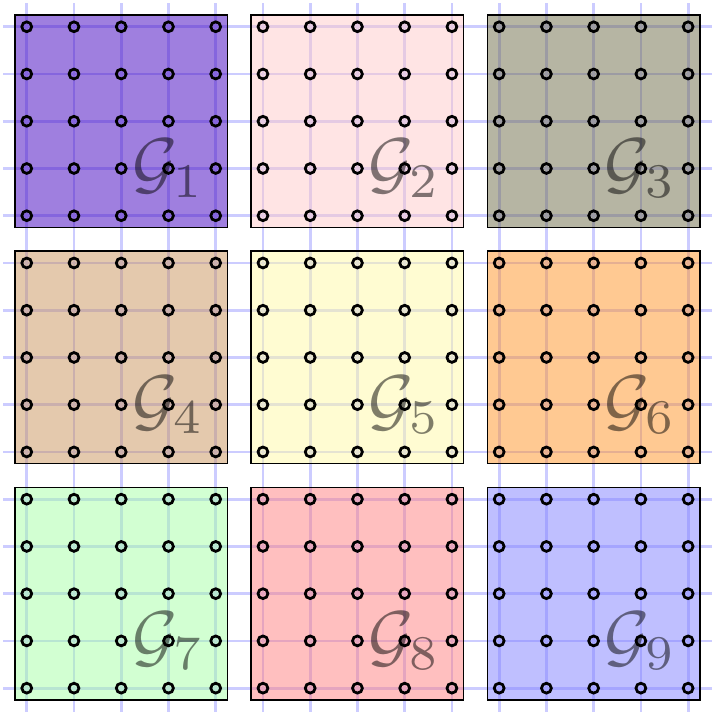}
\label{PatchPatch}}}
\caption{(a) Block of a hyperspectral image. Shaded areas show windows centered at pixels $x_i$ and $x_j$.  The contextual representation of a pixel is computed from pixels inside its window. (b) Pixels of a hyperspectral image partitioned into $5\times5$ image patches, shown as shaded square regions.}
\label{Context_image}
\end{figure}

%\begin{figure}[!t]
%\centering
%\subfloat{\includegraphics[width=1.7in]{Neib_S}}%
%\label{Neib_S}
%\hfil
%\subfloat{\includegraphics[width=1.7in]{PatchPatch}%
%\label{PatchPatch}}
%\caption{(a) Block of a hyperspectral image. Shaded areas show windows centered at pixels $x_i$ and $x_j$.  The contextual representation of a pixel is computed from pixels inside its window. (b) Pixels of a hyperspectral image partitioned into 9 contextual groups, shown as shaded square regions.}
%\label{Context_image}
%\end{figure}

As was mentioned at the beginning of Section \ref{DL_HSI}, the basic dictionary learning model treats $x_1,\ldots,x_N$ independently and is unable to capture contextual information unless it is provided explicitly in the training data.  To this end, let us define $c_1,\ldots,c_N$ as the contextual representation of the $N$ training pixels with respective spectral representations $x_1,\ldots,x_N$. Similar to \cite{Camps06}, we define $c_i$ as the moments computed separately over each spectral channel, from the samples surrounding $x_i$. Since the pixels surrounding $x_i$ are often from the same class, this is similar to computing an estimate of the class moments local to $x_i$. Extracting the first moment in a local manner may also be viewed as applying a lowpass filter to the image in order to remove noise and derive features that are locally more alike. Also, as discussed in \cite{Landgrebe03}, the first and second moments of a class are particularly informative for HSI classification. Therefore, we set $c_i$ to represent either the first moment or the concatenation of the first and second moments. Similar to \cite{Camps06,Camps10,Chen11}, the surrounding pixels of $x_i$ are defined as those inside a square centered at $x_i$ (See Fig. \ref{Neib_S}) the width of which is determined by cross validation. Once more, applying the dictionary learning formulation of (\ref{DL_Simple}) with $X$ replaced by $C=[c_1,\ldots,c_N]$, results in corresponding sparse representations $y_1,\ldots,y_N$ and the dictionary $D$, which may be used to train a linear SVM. Given a new spectral sample $x$, we calculate its contextual representation $c$, and apply sparse coding to find the corresponding sparse representation $y$, which is again classified using the trained linear SVM. We shall refer to the basic dictionary model applied to contextual (spectral) data for HSI classification as contextual (spectral) dictionary learning.

\subsection{Spectral-Contextual Dictionary Learning}\label{Define_Contextual}
%\begin{figure}[!t]
%\centering
%\includegraphics[width=1.5in]{PatchPatch}
%\caption{Pixels of a hyperspectral image partitioned into 9 contextual groups, shown as shaded square regions.}
%\label{PatchPatch}
%\end{figure}
Methods such as contextual dictionary learning, the composite kernel SVM of \cite{Camps06}, and the joint sparsity model of \cite{Chen11} that use a window centered at the pixel of interest to calculate contextual information, have two important drawbacks. First, as is shown in Fig. \ref{Neib_S} the windows that belong to neighboring pixels have large overlaps, which hinders the potential for parallel computation \cite{Plaza09}. Second, these methods gain their classification power largely from contextual information. In the aforementioned contextual dictionary learning or Chen et al.'s joint sparse model \cite{Chen11}, the spectral representation of the center pixel has as much significance in finding its label as any of the other pixels inside its window. Also, as we shall see in the experiments of Section \ref{Experiments}, the weighted sum of spectral and contextual kernels \cite{Camps06} achieves highest accuracy when the weight of the spectral kernel is zero.

The advantages of dictionary learning for HSI data modeling as explained in Section \ref{Intro}, together with the drawbacks of contextual-based methods as described above, and the limitations of the basic dictionary learning model, motivate the use of the structured dictionary learning model which can take advantage of both spectral and contextual information in HSI. To achieve this goal, we denote by $x_1,\ldots,x_N$ the spectral representation of the pixels in a hyperspectral image and define the contextual groups $\mathcal{G}_1,\ldots,\mathcal{G}_g$, as non-overlapping image patches. Each patch is a $w\times w$ square of pixels\footnote{If the image width or height are not divisible by $w$, at the bottom and right edges of the image the patches become smaller rectangles, }. Fig. \ref{PatchPatch} shows how the pixels of a hyperspectral image may be partitioned into $5\times5$ squares of pixels. A number of other ways to define $\mathcal{G}_1,\ldots,\mathcal{G}_g$ are imaginable. For example, HSI segmentation methods \cite{Plaza09,Tarabalka09} have a vast literature that may aid in defining more intelligent contextual groups that yield better results, yet we use the method illustrated in Fig. \ref{PatchPatch} for its simplicity and speed, and leave more complex methods to future research. In order to find the dictionary $D$ and sparse representations $y_1,\ldots,y_N$, we employ the dictionary learning formulation of (\ref{DL_L2GPD}).
%In order to find the dictionary $D$ and sparse representations $y_1,\ldots,y_N$ using the dictionary learning formulation of (\ref{DL_L2GPD}), $q_{_\mathcal{G}}$ and $\lambda_{\mathcal{G}}$ need to be defined. For (\ref{L2GPD}) to be a valid pdf, $q_{_\mathcal{G}}$ should be larger than $|\mathcal{G}|-1$. Hence, we set $q_{_\mathcal{G}}=|\mathcal{G}|$ and use $\lambda_\mathcal{G}=\sqrt{|\mathcal{G}|}$ to achieve similar sparsity in groups of different size. 
Once $y_1,\ldots,y_N$ are computed,
%from (\ref{DL_L2GPD}),
we train a linear SVM on the sparse representation of the training data and classify the sparse representation of the test samples.

\section{Experimental Results and Analysis}\label{Experiments}
In this section, we provide experimental results to validate the effectiveness of the proposed structured dictionary-based model for classifying real hyperspectral images. We compare the classification accuracies of the dictionary-based models, namely Spectral Dictionary Learning \cite{Charles11} (SDL), Contextual Dictionary Learning (CDL), and Spectral-Contextual Dictionary Learning (SCDL) with that of several methods. Support vector machines have proven successful in supervised classification of high-dimensional data such as HSI. Hence, we compare our results to SVM applied to spectral data with a polynomial kernel (SVM) \cite{Gualtieri99}, SVM applied to contextual data with an RBF kernel (CSVM) \cite{Camps06}, and SVM with a weighted sum kernel from  the composite kernel framework (CKSVM) \cite{Camps06}. Among the joint-sparsity-based HSI classification methods proposed recently in \cite{Chen11}, the Simultaneous Orthogonal Matching Pursuit (SOMP)  algorithm has achieved the best results, which we shall also use for comparison. We also compare the classification accuracy of SCDL with Dictionary Modeling with Spatial coherence (DMS) \cite{Castrodad11} and the graph kernel SVM (GKSVM) of \cite{Camps10} using the settings and results reported therein. In this section, methods that use contextual information collected from a window surrounding a pixel are further distinguished using $\mu$ for those that use the first moment, and $\mu,\sigma$ for those that employ both the first and second moments.

For the SVM-based approaches (SVM, CSVM, and CKSVM) all parameters (polynomial kernel degree $d$, RBF-kernel parameter $\sigma$, regularization parameter $C$, the composite kernel weight $\alpha$, and the window width $w$) are obtained by five-fold cross validation. Table \ref{Table_CV_Range} shows the values that each variable was allowed to take in cross validation. For SOMP the same values of window size and sparsity level as those reported in \cite{Chen11} were used. The SOMP algorithm uses the $\ell_p$ norm to iteratively find the best atoms from the dictionary. At each iteration the correlation of all the signal residuals inside the window are found with each atom, and the atom for which the $\ell_p$ norm of the correlation vector is largest is chosen. As explained in \cite{Chen11}, $p=1,2,\infty$ are most common. Unfortunately, the specific value of $p$ used for each dataset is not given. Hence, we have tested each of the three values and reported the accuracy using the value of $p$ that obtained the best results. For SOMP the complete set of training data was used as the dictionary. For SDL and CDL we also used five-fold cross validation to tune the parameters. The sparse regularization factor $\gamma$ in (\ref{DL_Simple}) was allowed to take values from $\{0.1,1,10,100\}$. To choose the number of dictionary atoms, values were chosen from $\{\frac{1}{2},\frac{1}{4},\frac{1}{8},\frac{1}{16}\}$, as a fraction of the number of training data. The window size for CDL was also chosen from the same range as CSVM and CKSVM shown in Table \ref{Table_CV_Range}. For SCDL we use $\sigma^2=10$ for all datasets to obtain $\gamma_{_\mathcal{G}}$ for each contextual group in (\ref{DL_L2GPD}). For the relatively small Aviris Indian Pines dataset with lower spatial resolution we used a random selection of $\frac{1}{2}$ training data to initialize the dictionary and an $8\times8$ patch size. For the relatively large ROSIS Urban datasets with high spatial resolution we used $\frac{1}{8}$ training data to initialize the dictionary and a larger $16\times16$ patch size. 

\begin{table}[!t]
\renewcommand{\arraystretch}{1.2}
 %if using array.sty, it might be a good idea to tweak the value of
 %\extrarowheight %as needed to properly center the text within the cells
\caption{Range of Values Used in Cross Validation for SVM, CSVM, and CKSVM}
\label{Table_CV_Range}
\centering
% Some packages, such as MDW tools, offer better commands for making tables
% than the plain LaTeX2e tabular which is used here.
\includegraphics{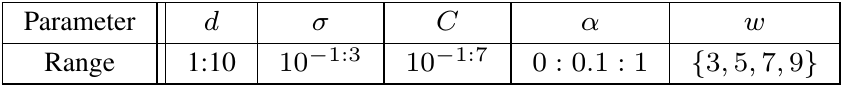}
%\begin{tabular}{|c||c|c|c|c|c|}
%\hline
%Parameter & $d$ & $\sigma$ & $C$ & $\mu$ & $w$\\
%\hline
%Range & 1:10 & $10^{-1:3}$ &  $10^{-1:7}$  & $0:0.1:1$ & $\{3,5,7,9\}$\\
%\hline
%\end{tabular}
\end{table}

Of the reported methods, some require an $M$-class SVM classification, be it linear or nonlinear. We use the one-against-one strategy for multi class classification using SVMs. That is, $\binom{M}{2}$ binary SVM classifiers are trained, one for each pair of classes. To classify a new test sample, it is applied to all classifiers and the label chosen by the majority of the classifiers is selected. For our experiments we used the LIBSVM \cite{LIBSVM11} implementation\footnote{All code used for the experiments is available at \url{http://ssp.dml.ir/wp-content/uploads/2013/07/HSI.zip}}.

\subsection{AVIRIS Indian Pines Dataset}
\begin{figure*}[!t]
\centering
\subfloat[]{\includegraphics[width=1.3in]{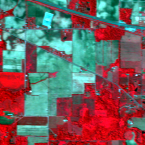}%
\label{IP_orig}}
\hfil
\subfloat[]{\includegraphics[width=1.3in]{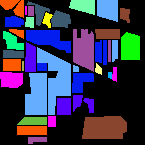}%
\label{IP_GT}}
\hfil
\subfloat[]{\includegraphics[width=1.3in]{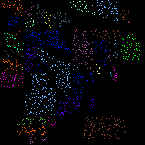}%
\label{IP_train}}
\hfil
\subfloat[]{\includegraphics[width=1.3in]{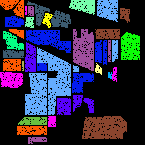}%
\label{IP_test}}
\hfil
\subfloat[]{\includegraphics[width=1.3in]{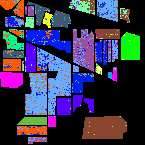}%
\label{IP_SVM}}
\\
\subfloat[]{\includegraphics[width=1.3in]{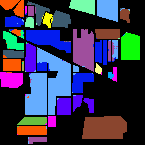}%
\label{IP_CSVM}}
\hfil
\subfloat[]{\includegraphics[width=1.3in]{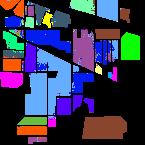}%
\label{IP_SOMP}}
\hfil
\subfloat[]{\includegraphics[width=1.3in]{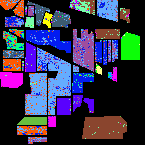}%
\label{IP_SDL}}
\hfil
\subfloat[]{\includegraphics[width=1.3in]{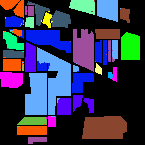}%
\label{IP_CDL}}
\hfil
\subfloat[]{\includegraphics[width=1.3in]{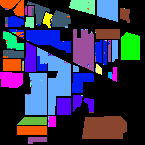}%
\label{IP_SCDL}}
\caption{Indian Pines image. (a) Three-band color composite (bands 50,27, and 17). (b) Ground-truth. (c) Training data. (d) Test data. Classification maps obtained by (e) SVM, (f) CSVM$_\mu$, (g) SOMP, (h) SDL, (i) CDL$_\mu$, and (j) SCDL colored according to Table \ref{Indian_Pines_Classes}.}
\label{IP_image}
\end{figure*}

One of the datasets that is often used for evaluating HSI classification is the Indian Pines image \cite{Aviris}. It was collected over an agricultural/forested area in NW Indiana using the AVIRIS sensor. The image is $145\times145$ pixels in size with a spatial resolution of 20 m/pixel and consists of 16 ground-truth classes. The spectral vectors consist of 220 bands across the spectral range 0.2 to 2.4 $\mu$m, of which 20 noisy bands (104-108, 150-163, 220) corresponding to the region of water absorption are removed. Fig. \ref{IP_image} shows a color composite image of the Indian Pines dataset along with the ground-truth. Around 10\% of the data are randomly chosen for training and the remaining 90\% are used for testing. The specific classes and the number of train and test data in each class are reported in Table \ref{Indian_Pines_Classes}.

\begin{table}[!t]
\renewcommand{\arraystretch}{1}
 %if using array.sty, it might be a good idea to tweak the value of
 %\extrarowheight %as needed to properly center the text within the cells
\caption{Indian Pines Ground-Truth Classes and Train/Test Sets }
\label{Indian_Pines_Classes}
\centering
% Some packages, such as MDW tools, offer better commands for making tables
% than the plain LaTeX2e tabular which is used here.
\includegraphics{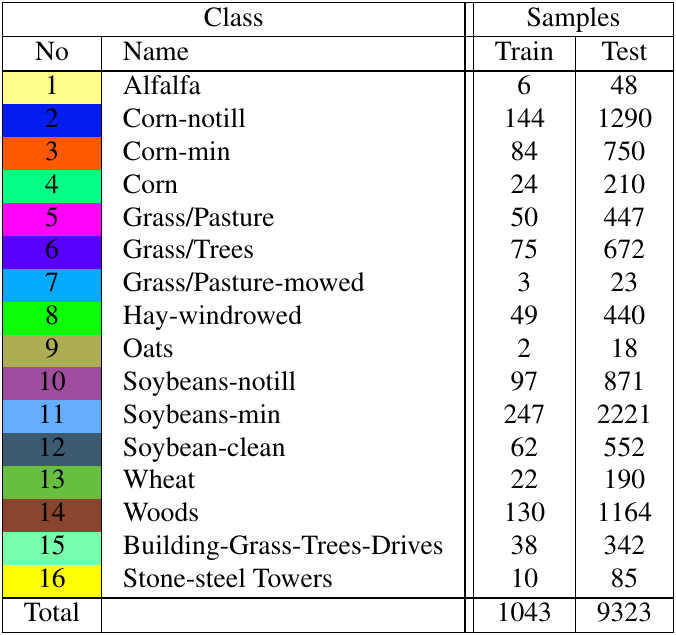}
%\begin{tabular}{|c|l||c|c|}
%\hline
%\multicolumn{2}{|c||}{Class} & \multicolumn{2}{c|}{Samples} \\
%\hline
%No & Name & Train & Test\\
%\hline
%1 \cellcolor{IP1}& Alfalfa & 6 & 48 \\
%2 \cellcolor{IP2}& Corn-notill & 144 & 1290 \\
%3 \cellcolor{IP3}& Corn-min & 84 & 750 \\
%4 \cellcolor{IP4}& Corn & 24 & 210 \\
%5 \cellcolor{IP5}& Grass/Pasture & 50 & 447 \\
%6 \cellcolor{IP6}& Grass/Trees & 75 & 672 \\
%7 \cellcolor{IP7}& Grass/Pasture-mowed & 3 & 23 \\
%8 \cellcolor{IP8}& Hay-windrowed & 49 & 440 \\
%9 \cellcolor{IP9}& Oats & 2 & 18 \\
%10 \cellcolor{IP10}& Soybeans-notill & 97 & 871 \\
%11 \cellcolor{IP11}& Soybeans-min & 247 & 2221 \\
%12 \cellcolor{IP12}& Soybean-clean & 62 & 552 \\
%13 \cellcolor{IP13}& Wheat & 22 & 190 \\
%14 \cellcolor{IP14}& Woods & 130 & 1164 \\
%15 \cellcolor{IP15}& Building-Grass-Trees-Drives & 38 & 342 \\
%16 \cellcolor{IP16}& Stone-steel Towers & 10 & 85 \\
%\hline
%Total & & 1043 & 9323\\
%\hline\end{tabular}
\end{table}

%\begin{figure}[!t]
%\centering
%\subfloat[]{\includegraphics[width=1.4in]{IP_orig.png}%
%\label{IP_orig}}
%\hfil
%\subfloat[]{\includegraphics[width=1.4in]{IP_GT.png}%
%\label{IP_GT}}
%\caption{Indian Pines image. (a) Three-band color composite (bands 50,27, and 17). (b) Ground-truth colored according to Table \ref{Indian_Pines_Classes}.}
%\label{IP_image}
%\end{figure}

\begin{table*}[!t]
\renewcommand{\arraystretch}{1}
 %if using array.sty, it might be a good idea to tweak the value of
 %\extrarowheight %as needed to properly center the text within the cells
\caption{Classification Accuracy (\%) for AVIRIS Indian Pines for Different Classifiers}
\label{Indian_Pines_Accuracy}
\centering
% Some packages, such as MDW tools, offer better commands for making tables
% than the plain LaTeX2e tabular which is used here.
\includegraphics{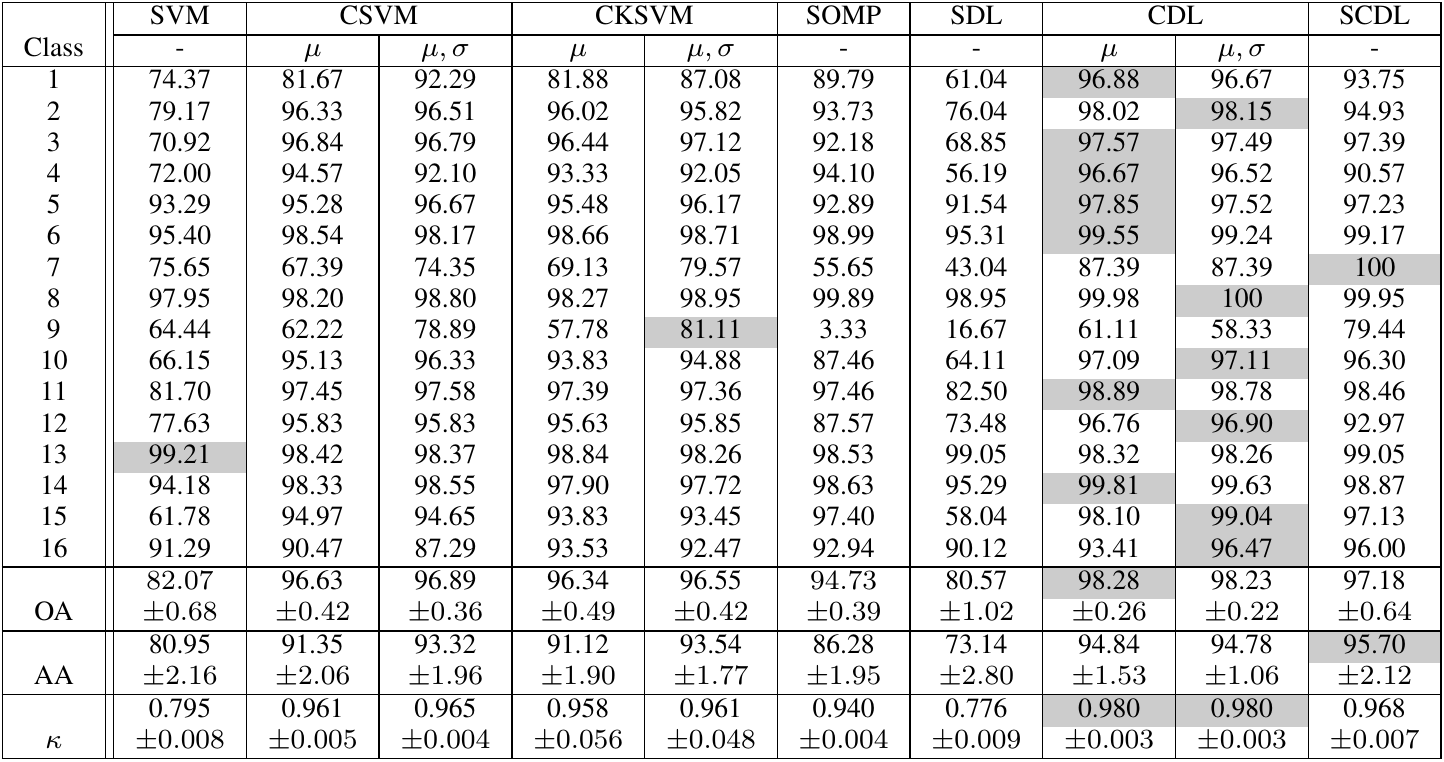}
\end{table*}

The different classification approaches are compared in Table \ref{Indian_Pines_Accuracy}, where the classification accuracy for each class, overall accuracy (OA), average accuracy (AA), and the $\kappa$ coefficient measure \cite{Richards06} are reported. The overall accuracy measures the ratio of correctly classified pixels to all test pixels, while the average accuracy is simply the average of the accuracies for each class. The $\kappa$ coefficient is computed from different entries in the confusion matrix and is a robust measure of agreement that corrects for random classification. The results are averaged over ten runs and the standard deviation is also reported for OA, AA, and $\kappa$. The test and train data of the first run and the obtained classification map for each method is depicted in Fig. \ref{IP_image}. There are a few observations to be made here. First, the methods based only on spectral characteristics (SVM and SDL) provide poor accuracies compared to the other methods that take into account contextual information. This stresses yet again the importance of contextual data for HSI classification. Second, the contextual kernel SVM (CSVM) provides marginally better results than the composite kernel SVM (CKSVM), which shows that CKSVM is unsuccessful in combining spectral and contextual characteristics of HSI. This is not very surprising considering the accuracies reported for SVM and CSVM which yields the spectral kernel unlikely to contain complementary information. Third, the SOMP approach provides results comparable to CSVM for $p=\infty$, which means that the training data inside the window surrounding a pixel are always chosen as part of the best atoms, since each is both present in the window and the dictionary, and has maximum correlation with itself. This is somewhat similar to a $k$-nearest neighbor approach. This is also why the method obtains poor results for Oat pixels which cover a narrow region. As noted by Chen et. al \cite{Chen11}, the local window for an Oat pixel is dominated by pixels from two adjacent classes. Fourth, since the Indian Pines map contains very large homogenous regions (Fig. \ref{IP_GT}) there is little difference between using the first moment or using both the first and second moments inside a window as the contextual representation of a pixel. Finally, whilst CDL provides the best overall accuracy, its average accuracy is slightly lower than that of SCDL. This is due to the fact that SCDL is able to employ both spectral and contextual information, which puts CDL in a disadvantage when it comes to classifying classes with narrow regions (classes 7 and 9). SCDL provides high accuracies for each class but trails slightly behind CDL due to the mostly large homogenous regions present in the Indian Pines image. This shows that SCDL is likely to perform better than CDL for urban scenery where there are few large homogenous areas, as we see in the next section.

\begin{figure}[!t]
\centering
\includegraphics[width=3.75in]{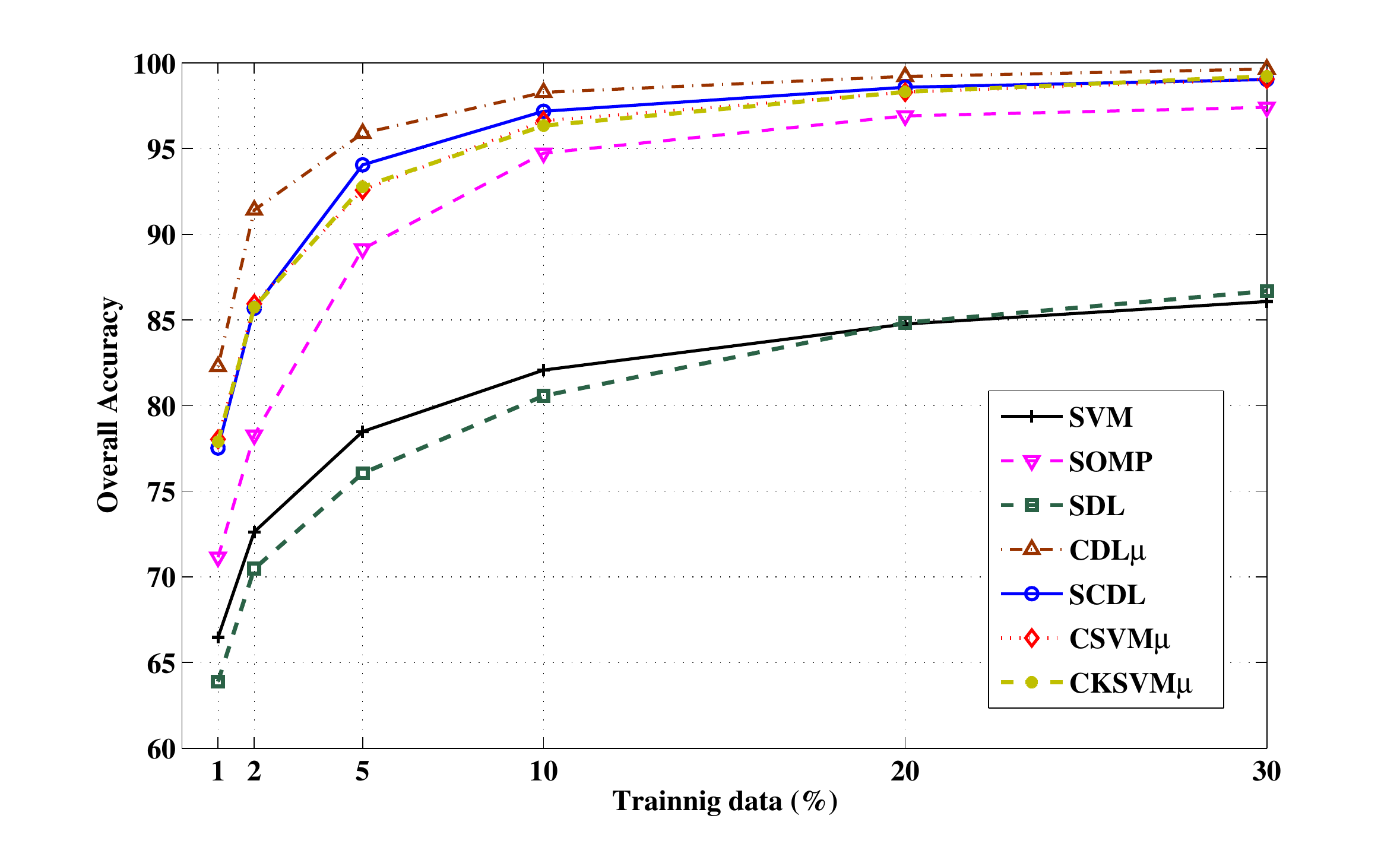}
\caption{Effect of number of training data on classification results for Indian Pines.}
\label{Method_Effect}
\end{figure}

To further compare the different classification approaches reported in Table \ref{Indian_Pines_Accuracy}, we have tested them on different training sample sizes. Fig. \ref{Method_Effect} shows the overall accuracy of each method for 1, 2, 5, 10, 20 and 30\% training data averaged over ten runs. To avoid crowding the figure we have left out methods that use both the first and second moments since they have very similar results to the corresponding method that uses only the first moment. As the results show, CDL provides a large improvement for small number of training samples with SCDL trailing slightly behind. As was mentioned in Section \ref{Intro}, dictionary learning has proven successful in classification problems where few training samples are available. As was expected, SOMP falls further behind as the number of training samples is decreased due to its similarity to a $k$-nearest neighbor approach.

For completeness, we also compare SCDL with the GKSVM \cite{Camps10} and DMS \cite{Castrodad11} using the same settings reported there. For 1\% training data of the Indian Pines image, the authors reported 72.3\% and 73.4\% overall classification accuracy for GKSVM and DMS respectively. We obtained 76.2\% using SCDL. For 5\% training data the results reported in \cite{Castrodad11} are 83.5\% and 84.4\% for GKSVM and DMS respectively. Under this setting SCDL performs with 93.4\% overall classification accuracy. For these experiments the GKSVM uses an $11\times11$ window centered at the pixels of interest to gather contextual information at two scales, while DMS encourages smooth variation in the sparse representations for the four spatially connected neighbors, and as we mentioned earlier $8\times8$ contextual groups were employed for SCDL. Increasing the neighborhood size for DMS may increase the reported accuracy but would also incur a large increase in computational costs. It is also worth noting that the dictionary learned by SCDL has more than 50\% less dictionary atoms than DMS. To be fair, we also performed experiments with a $3\times3$ patch size for SCDL leading to 72.6\% and 87.9\% overall accuracy for 1\% and 5\% training data respectively.

\subsection{Dictionary Atoms and Sparse Representations}

\begin{figure*}[!t]
\centering
\includegraphics[width=6.in]{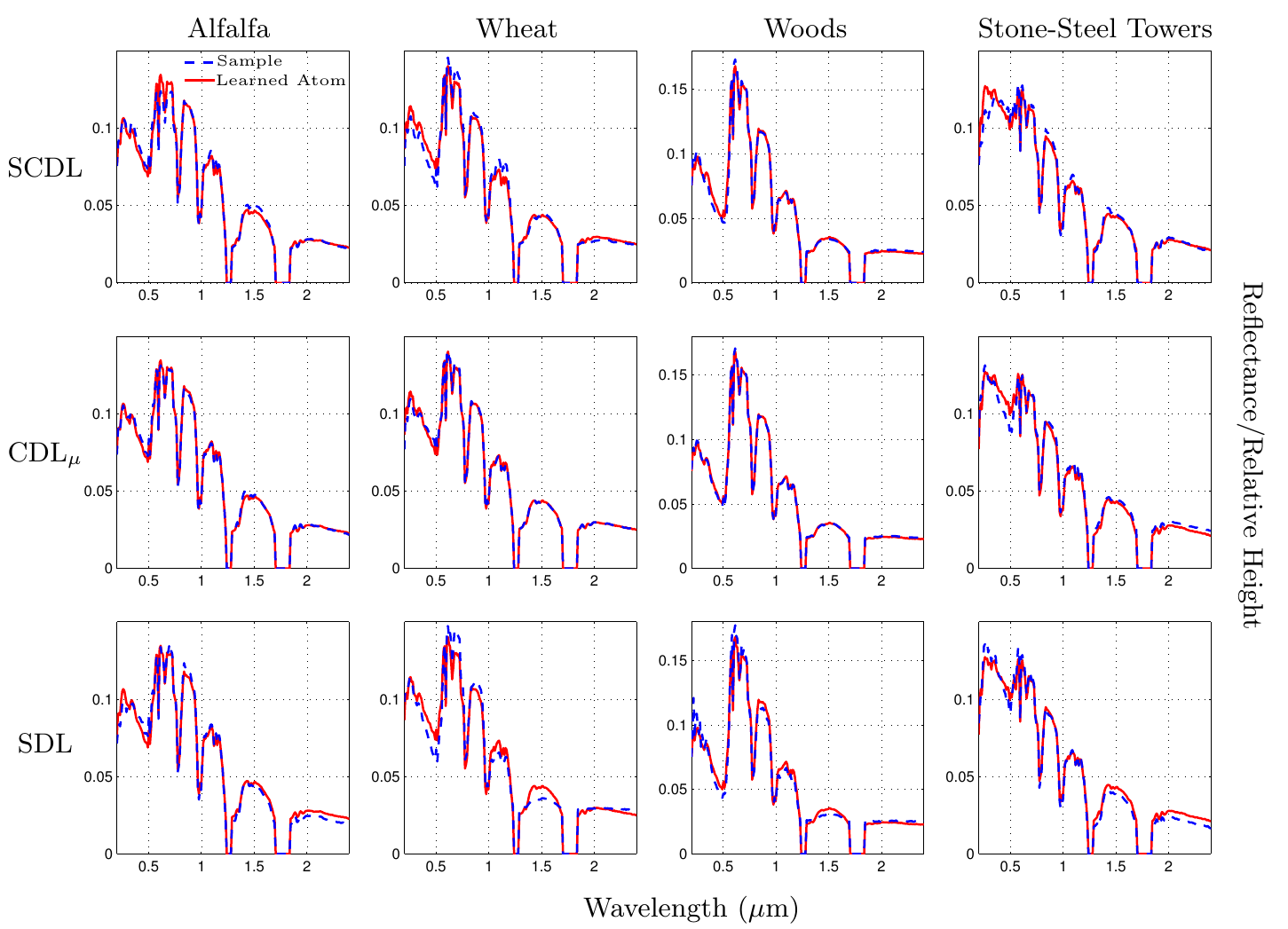}
\caption{Sample spectra for Alfalfa, Wheat, Woods, and Stone-Steel Towers in the Indian Pines dataset and the learned dictionary atom obtained by SDL, CDL, and SCDL that is closest to each sample. The two obvious gaps in the spectra correspond to the regions of water absorption which were removed.}
\label{Atoms}
\end{figure*}

\begin{figure*}[!t]
\centerline{\subfloat[]{\includegraphics[width =2.1in]{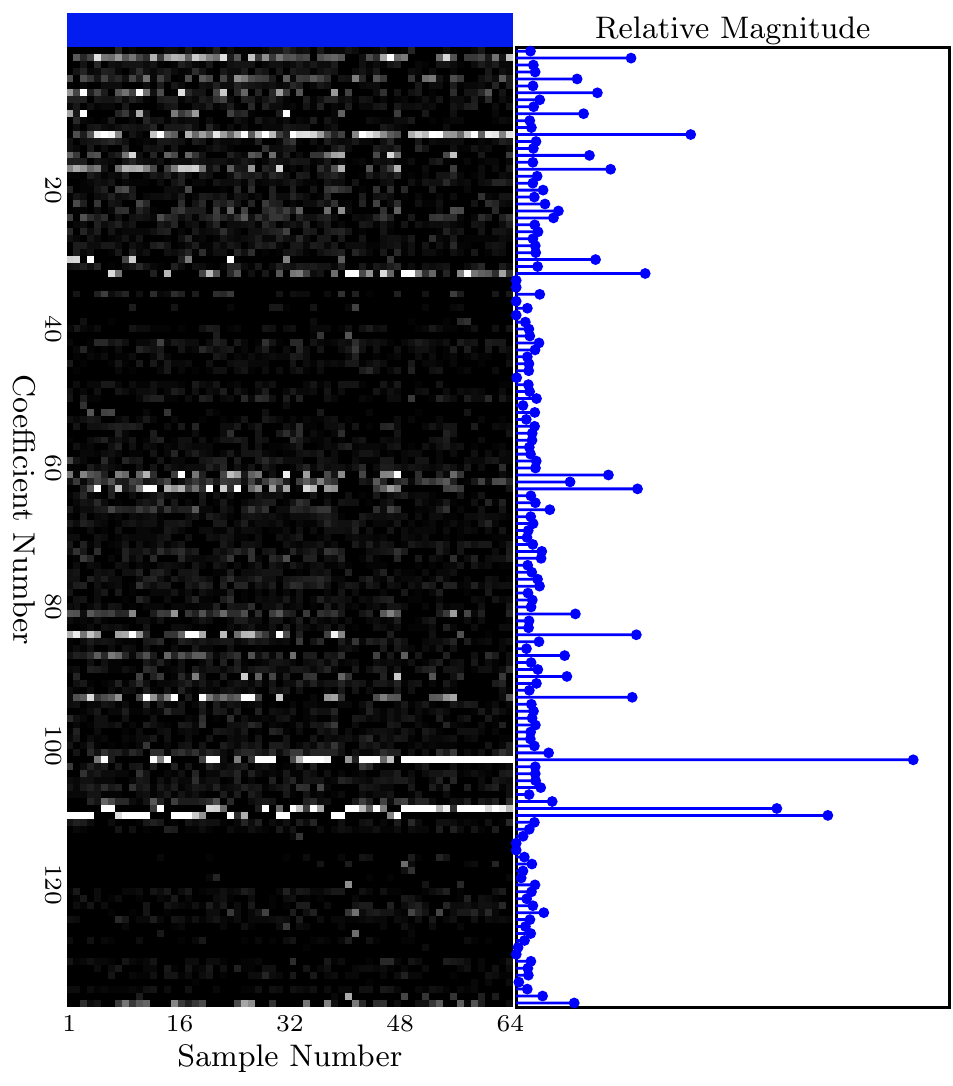}
\label{box_SDL}}
\hfil
\subfloat[]{\includegraphics[width=2.1in]{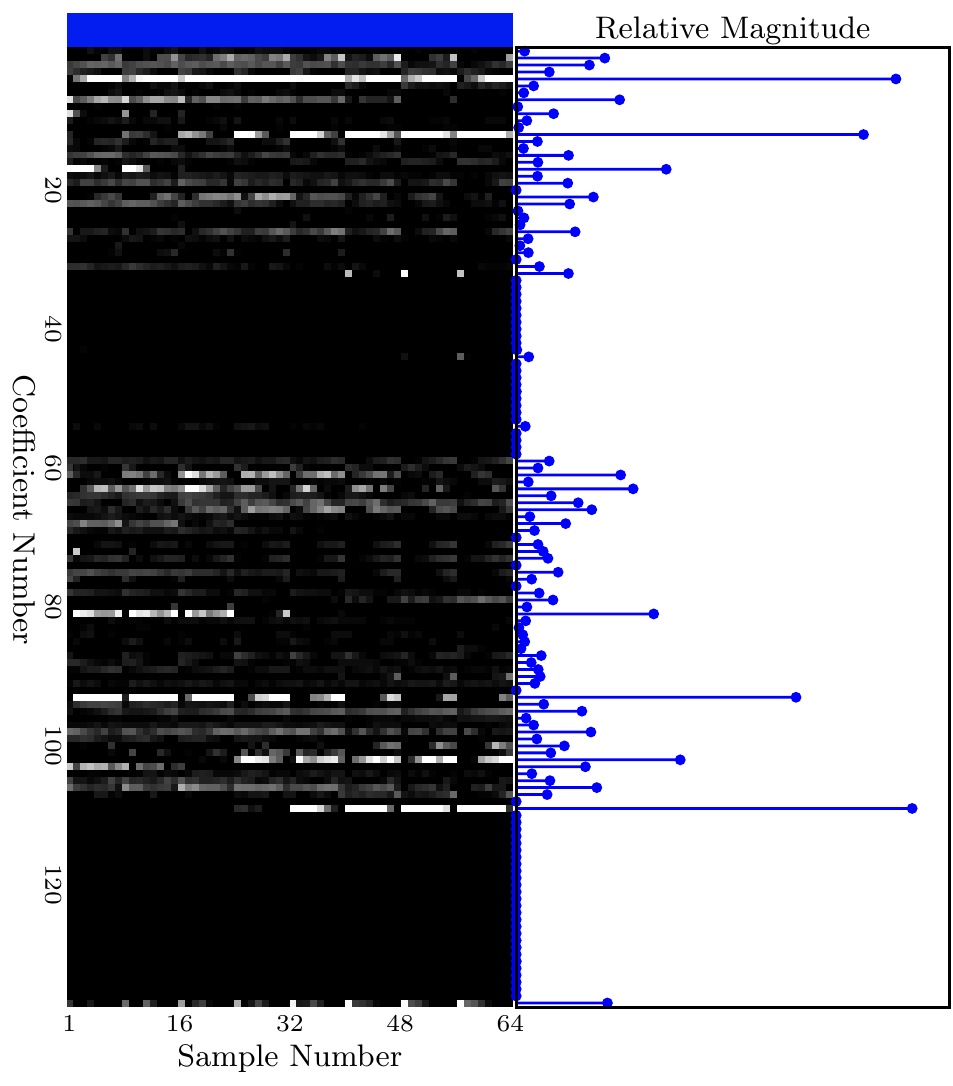}
\label{box_CDL}}
\hfil
\subfloat[]{\includegraphics[width=2.1in]{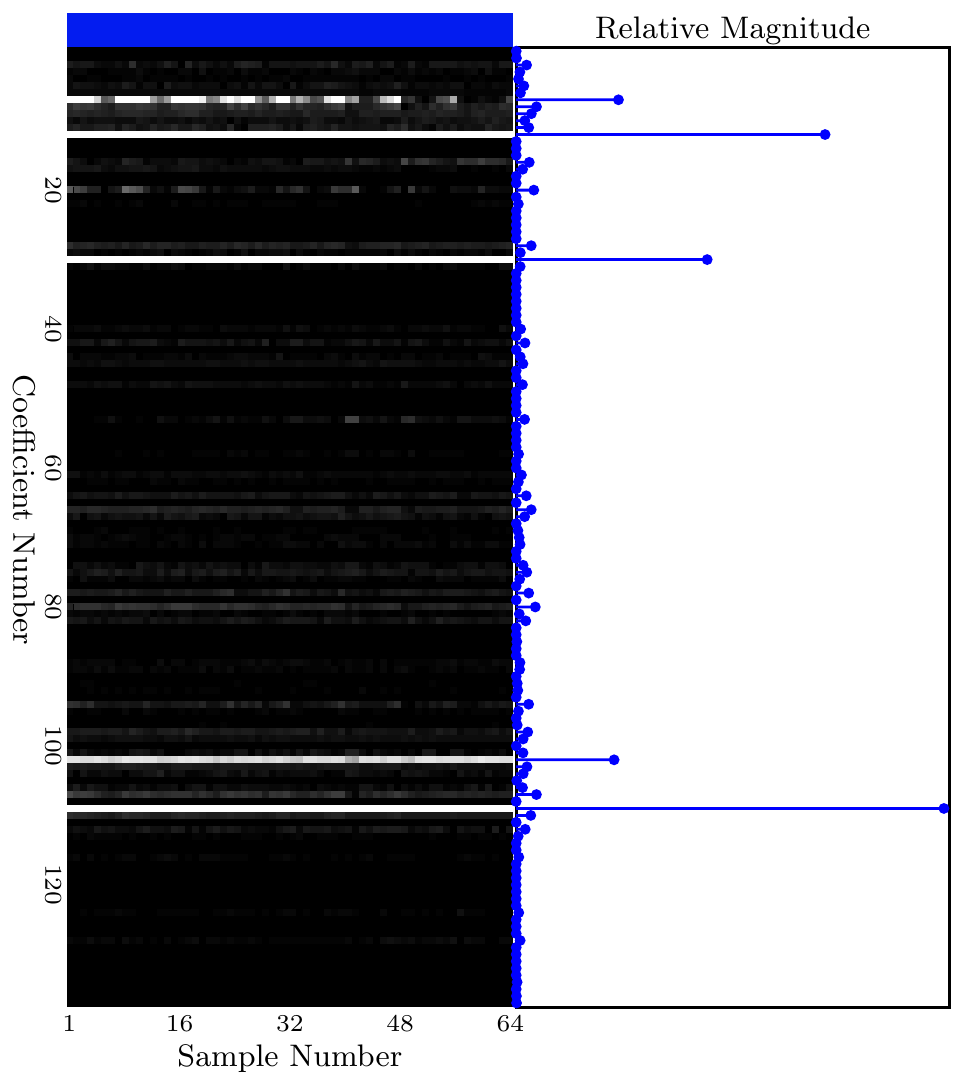}
\label{box_SCDL}}}
\caption{Sparse coefficients corresponding to samples inside an $8\times8$ contextual group of the Indian Pines image accompanied on the right by the norm of each row of the coefficient matrix for (a) SDL, (b) CDL, and (c) SCDL. The label of each sample is shown above its coefficients.}
\label{box}
\end{figure*}

\begin{figure*}[!t]
\centerline{\subfloat[]{\includegraphics[width =2.1in]{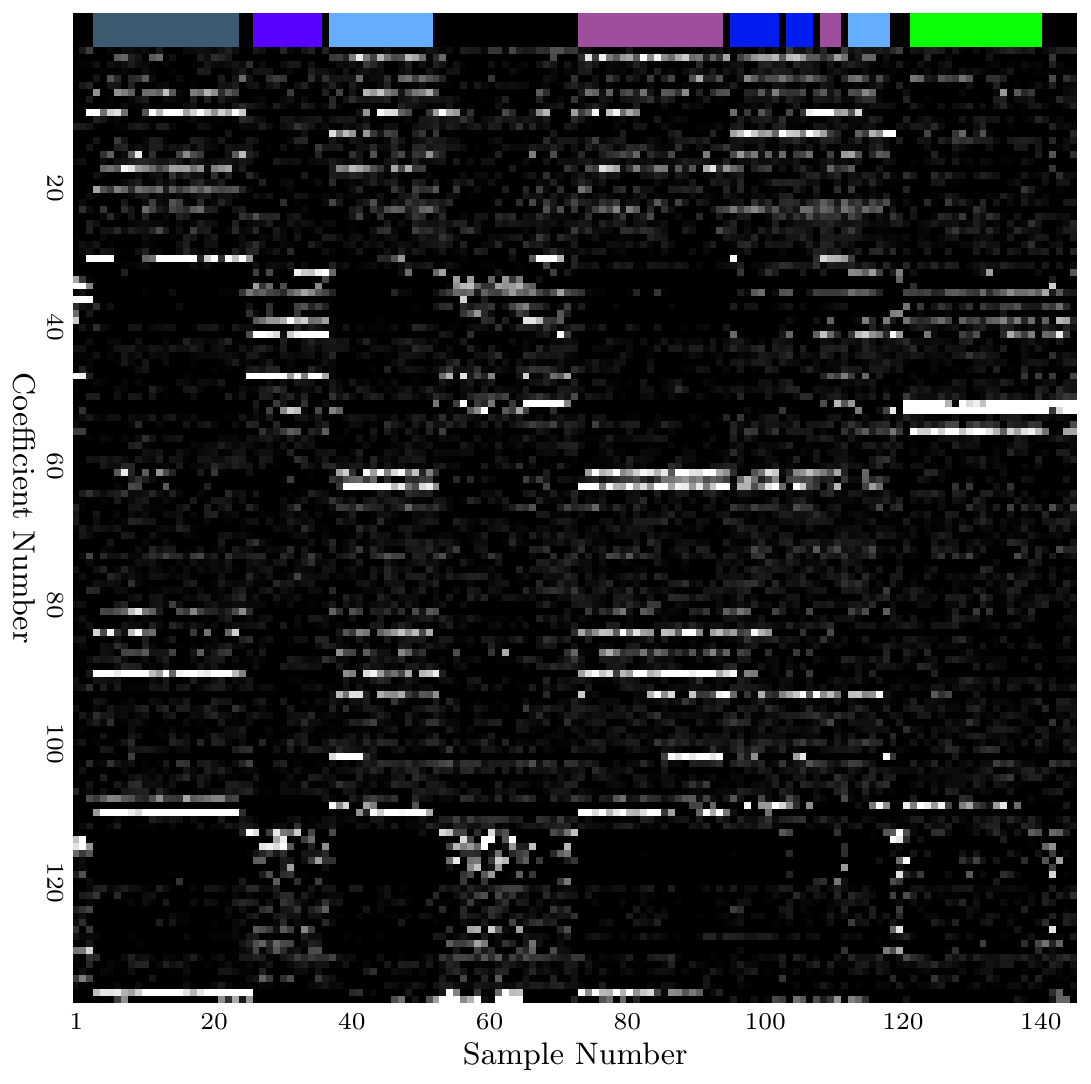}
\label{line_SDL}}
\hfil
\subfloat[]{\includegraphics[width=2.1in]{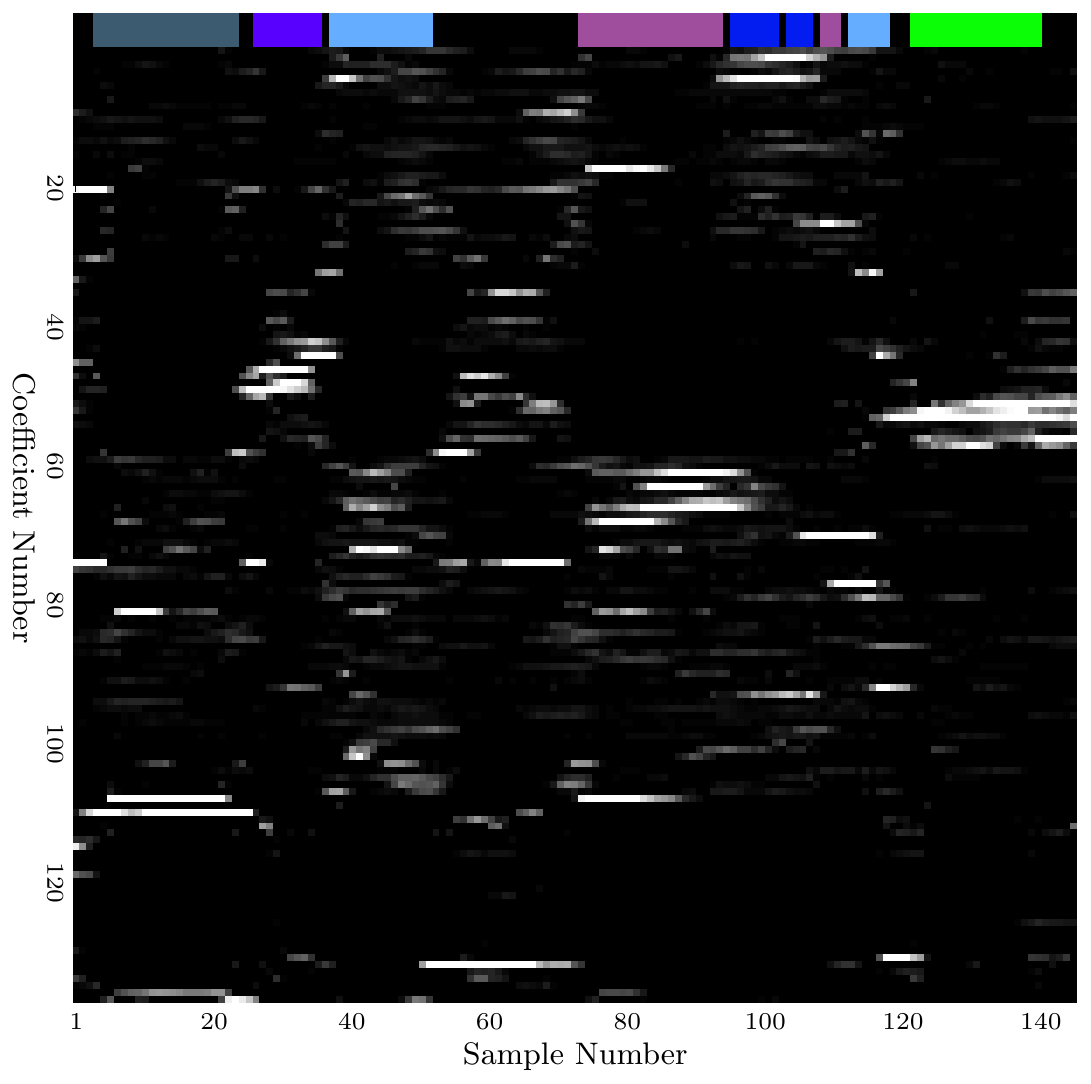}
\label{line_CDL}}
\hfil
\subfloat[]{\includegraphics[width=2.1in]{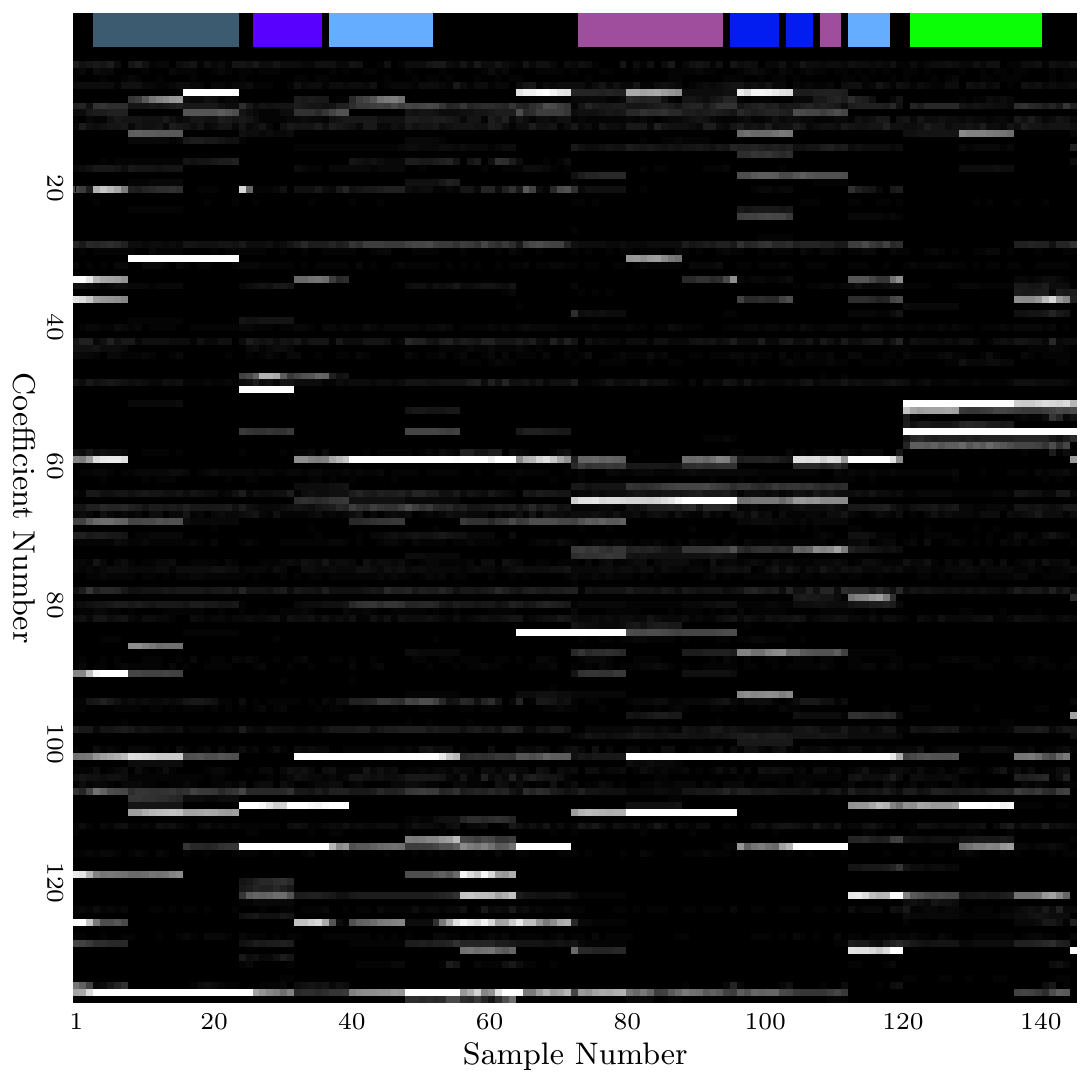}
\label{line_SCDL}}}
\caption{Sparse coefficients corresponding to 145 samples on a horizontal line of the Indian Pines image for (a) SDL, (b) CDL, and (c) SCDL. The label of each sample is shown above its coefficients.}
\label{line}
\end{figure*}

As a means of visual comparison, using 10\% of the Indian Pines training data, we learned dictionaries with 138 atoms ($\frac{1}{8}$th training data), using SDL, CDL$_\mu$, and SCDL. Fig. \ref{Atoms} depicts sample spectra for the classes Alfalfa, Wheat, Woods, and Stone-Steel Towers and the learned dictionary atom that is closest to each sample. An encouraging observation was that although we did not explicitly constrain neither the dictionary atoms nor the sparse representations to be nonnegative as in SDL, this was in fact almost always the case for CDL$_\mu$ and SCDL. The figure shows that all methods learn dictionary atoms that are quite similar to sample spectra obtained from the scene. This is inline with the observations made in \cite{Charles11}. CDL$_\mu$ seems to be more accurate in this regard, perhaps due to the denoising effect of spatial averaging. Since the dictionary atoms are similar to the sample spectra, the sparse representation of a sample with these atoms is likely to exhibit discriminative capabilities. To gain further insight, we have depicted in Fig. \ref{box} and Fig. \ref{line} the sparse representations obtained for an $8\times8$ contextual group, and also a line of 145 pixels in the Indian Pines dataset. From Fig. \ref{box_SCDL} one can see that a far fewer number of atoms are activated inside a contextual group for SCDL than for SDL (Fig. \ref{box_SDL}) or CDL$_\mu$ (Fig. \ref{box_CDL}). This was also observed in \cite{Iordache12_1} where all samples in an image were grouped together to decrease the total number of atoms activated. In Fig. \ref{line_SCDL} one may observe that for almost every instance in a class of data a distinctive atom is active. This is sometimes violated at the edges of a class, where instances from different classes fall into the same contextual group. Although, the edges are where most misclassifications occur (see Fig. \ref{IP_SCDL} and Fig. \ref{UoP_SCDL}), this is not always the case. The reason for this is that, while members of a contextual group are constrained to belong to the same subspace, they are not constrained to be any more similar within that subspace.

\subsection{Classifying at MSI-Resolution Via HSI-resolution Atoms}
A remarkable finding of \cite{Charles11} was that the sparse coding model accompanied by a dictionary learned from HSI-resolution data may be used to infer HSI-resolution spectra from simulated MSI-level measurements. Multispectral data generally have a much coarser spectral resolution (3-10 bands) than HSI data, and may be acquired in a more resource efficient manner. This motivates the need for processing at MSI-level with HSI-level quality. We follow the same experimental setting as \cite{Charles11} on the Indian Pines image, but since our goal is classification, we compare the overall classification accuracy using the retrieved sparse representations for SDL, CDL$_\mu$, and SCDL.
%To gain further insight, we also compare the ability of each method to infer sparse representations from MSI-level spectra that are similar to those inferred at HSI resolution. 

To simulate MSI-level resolution spectra, it is assumed that each band is a linear combination of some adjacent spectral bands in the original HSI data. To be specific MSI-level data, $z\in\mathbb{R}^b$ are simulated from HSI spectra via $z = Bx$, where $B\in\mathbb{R}^{b\times N}$ ($b<N$) sums adjacent spectral bands of $b$ non-overlapping bins in $x$. For one set of experiments $b=8$ bins are equally spaced in the lower half of the spectrum roughly corresponding to measurements obtained with the Worldview II MSI satellite \cite{Worldview} (MSI measurements), while for the second set of experiments, the $b=8$ bins cover the complete spectrum (coarse HSI measurements).

Given a pre-learned HSI-resolution dictionary $D$, using (\ref{LMM_DL}) with the same line of arguments as in Sections \ref{DL_SimpleSetting} and \ref{DL_HSI}, we obtain the MAP estimate for the sparse representations of SDL/CDL and SCDL as:
\begin{equation}\label{DL_SDL_MSI}
\arg\min_{Y}\frac{1}{2}\|Z-BDY\|_F^2 +\gamma\sum_{i=1}^N \|y_i\|_1
\end{equation}
and
\begin{equation}\label{DL_L2GPD_MSI}
\arg\min_{Y}\frac{1}{2} \|Z-BDY\|_F^2 + \sum_{i=1}^g \gamma_{_{\mathcal{G}_i}}\left\|Y_{\mathcal{G}_i}\right\|_{2,1}
\end{equation}
respectively, where $Z=[z_1,\ldots,z_N]$ are the MSI-resolution samples and $Y$ are constrained to be nonnegative for SDL.

\begin{figure}[!t]
\centering
\includegraphics[width=3.75in]{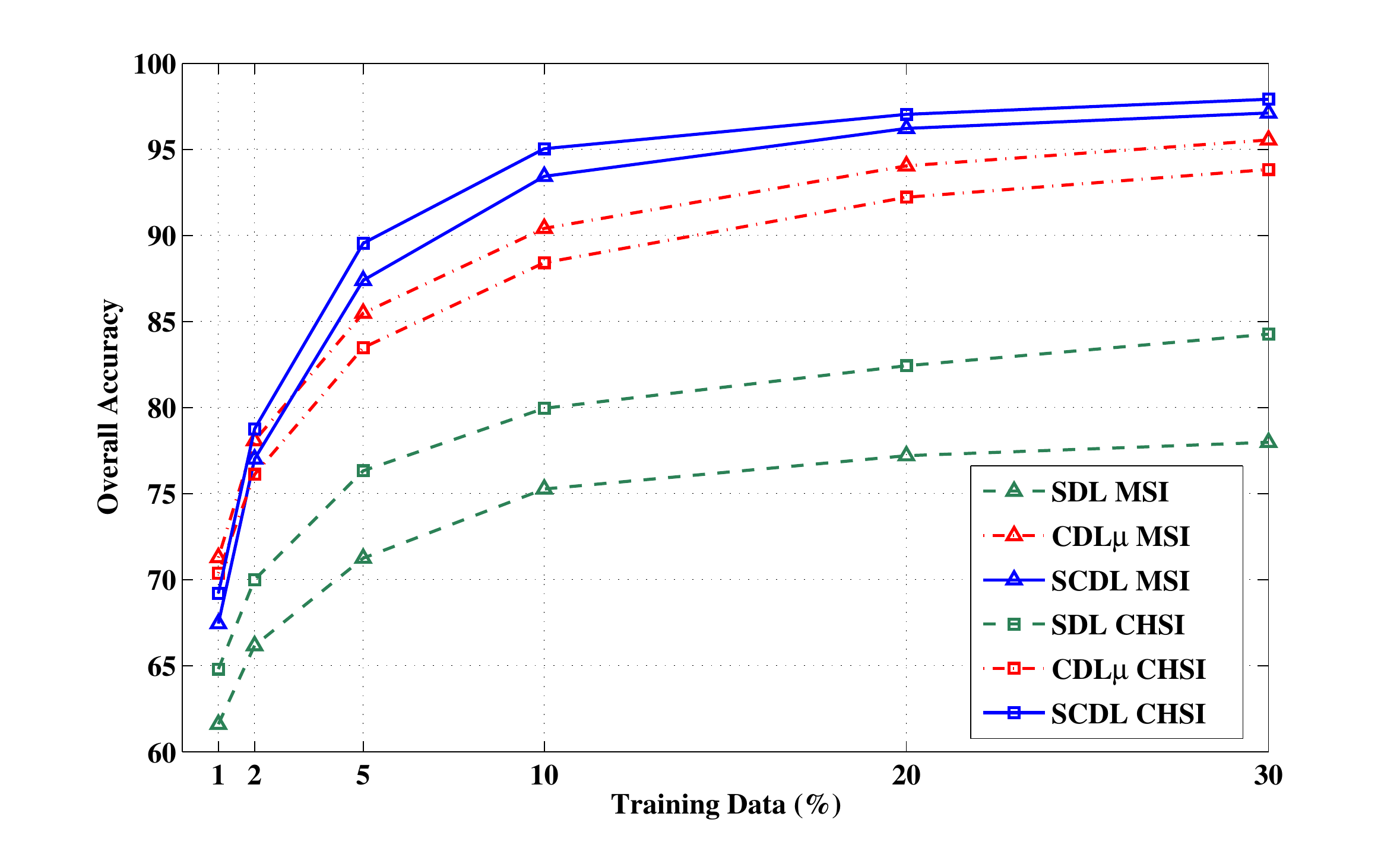}
\caption{Results for classification of the Indian Pines (synthetic) MSI and coarse HSI data using an HSI dictionary for different number of training data.}
\label{MSI_Effect}
\end{figure}

For each method, the samples are partitioned into initial train and test sets and the dictionary is learned. The test samples are then converted to MSI or coarse HSI (cHSI) measurements and again divided into test and train sets. After the sparse representations are obtained from (\ref{DL_SDL_MSI}) or (\ref{DL_L2GPD_MSI}), the train set is used to train a linear SVM and classification results are obtained for the test set. Fig. \ref{MSI_Effect} shows the overall classification accuracy obtained for each method with 1, 2, 5, 10, 20, and 30\% training data of the Indian Pines. The results show that all methods are able to obtain accuracies that are only slightly lower than those obtained with HSI-level measurements (Fig. \ref{Method_Effect}). For CDL the results have decreased more significantly than SCDL. Considering the largely coherent columns of $BD$ may explain this effect.
%SDL and CDL rely on only one pixel to obtain the corresponding representation
As the columns of $BD$ become more coherent, the sparse representations obtained by SDL or CDL exhibit larger spatial variations. This is not the case for SCDL which constrains pixels inside a contextual group to have a common sparsity pattern. We should note that these results serve as a proof of concept. In a more realistic scenario, the dictionary would be learned from a different but statistically similar scene, or from the same scene but at a different time.
%\begin{table}[!t]
%% increase table row spacing, adjust to taste
%\renewcommand{\arraystretch}{1}
% %if using array.sty, it might be a good idea to tweak the value of
% %\extrarowheight %as needed to properly center the text within the cells
%\caption{Spectral Recovery Error (SRE) and Coefficient Recovery Index (CRI) for (synthetic) MSI and coarse HSI measurements on the Indian Pines test set (10\% training data).}
%\label{MSI_RE}
%\centering
%% Some packages, such as MDW tools, offer better commands for making tables
%% than the plain LaTeX2e tabular which is used here.
%\includegraphics[width=3.45in]{Table_IX}
%\end{table}

\subsection{Parallel Processing of Contextual Groups}
In this section, we demonstrate how the computation cost of SCDL scales when contextual groups are processed in parallel with 1, 2 or 4 processing threads. We learned dictionaries initialized with $\frac{1}{2}$, $\frac{1}{4}$, and $\frac{1}{8}$ of the 10\% training data of the Indian Pines dataset with an $8\times8$ patch size and $\sigma^2=10$. The dictionary learning algorithm was permitted to continue for 100 iterations and the overall classification accuracy was recorded after each iteration. Experiments were performed on a machine with an Intel Core i5-3570K 64 bit processor and 16 GBs of RAM. The code for M-FOCUSS was written in C++ and all timings were performed using MATLAB.

\begin{figure}[!t]
\centering
\includegraphics[width=3.5in]{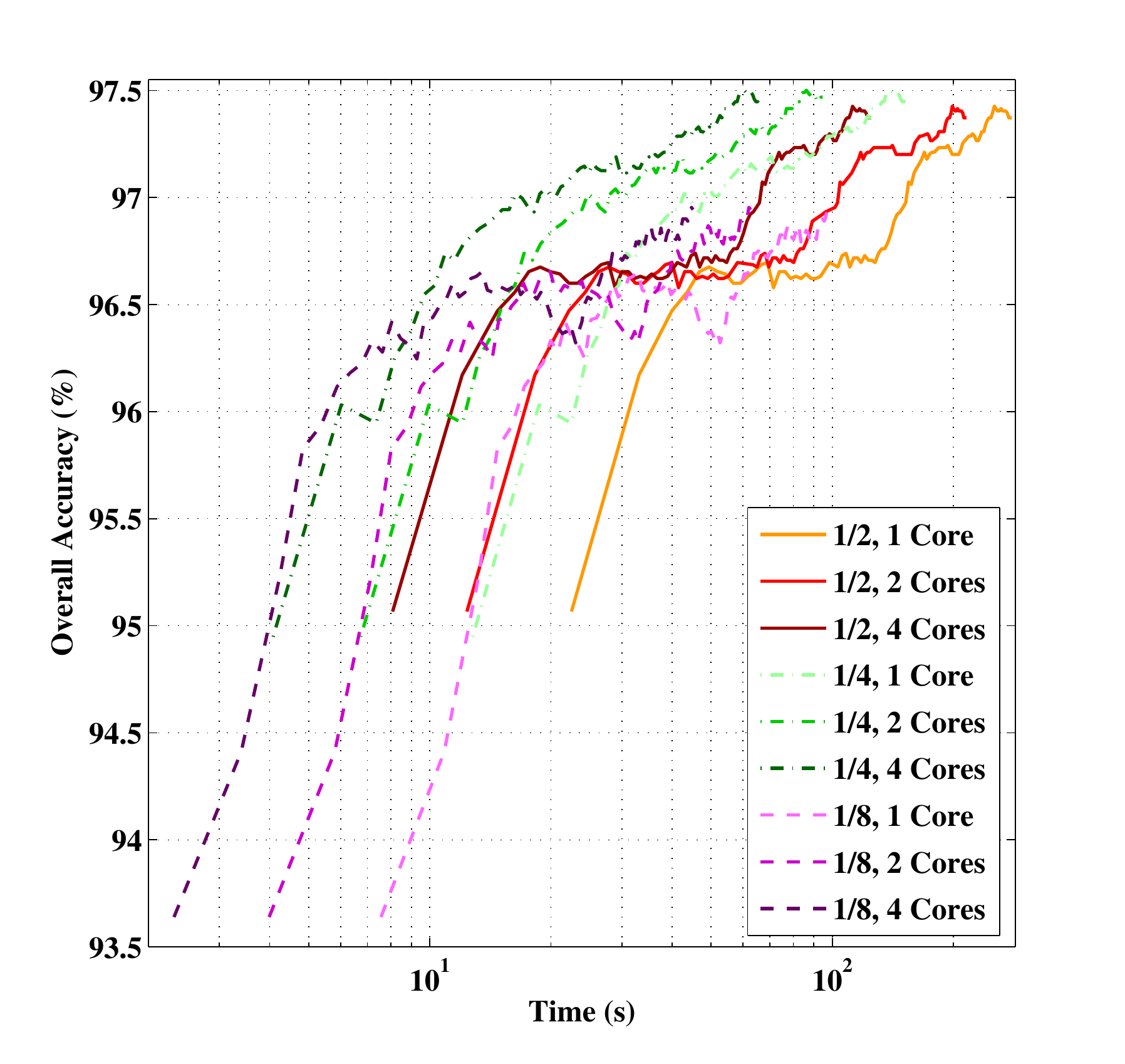}
\caption{Results for classification of the Indian Pines dataset with different dictionary size and different number of processing cores, as a function of time.}
\label{Parallel}
\end{figure}

Fig. \ref{Parallel} shows the overall classification accuracy on this dataset as a function of time. A few observations may be made. First, since the dictionary is initialized with training data from the scene, one can see how consecutive iterations of the learning process affect the discriminative ability of the sparse codes. Although, since the learning process is unsupervised the classification accuracy is not always increasing, overall, dictionary learning increases the discriminative ability of the codes as it adapts to the statistics of the scene. Second, in terms of classification accuracy, the size of the dictionary seems to matter most in the early iterations, with only a 0.5\% gap at the end for dictionaries initialized with $\frac{1}{8}$ and $\frac{1}{4}$ of the training data, and nearly no gap between $\frac{1}{4}$ and $\frac{1}{2}$. Third, the speedup gained from parallel processing is quite significant considering the different threads share the RAM. The first few iterations are often longer, while later iterations are shorter in time and thus are more affected by time required for thread initialization. Also, the algorithm scales quite well with the dictionary size, considering the dictionary is shared between parallel jobs. The speed up was around 1.8 for 2 threads independent of dictionary size and was 3.3, 3.1, and 2.8 as dictionary size is increased when 4 threads were used.

\subsection{ROSIS Urban Data over Pavia, Italy}
\begin{figure*}[!t]
\centering
\subfloat[]{\includegraphics[width=1.1in]{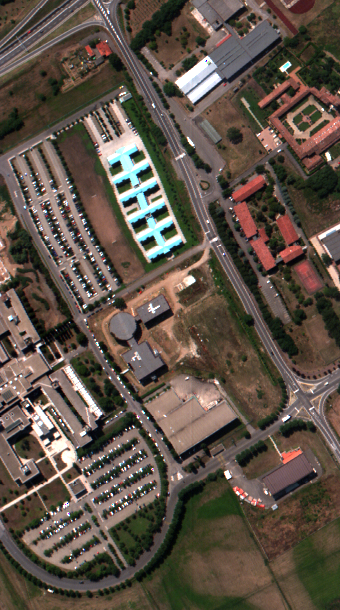}%
\label{UoP_orig}}
\hfil
\subfloat[]{\includegraphics[width=1.1in]{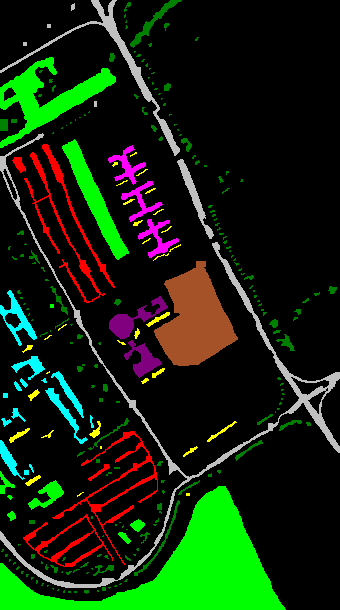}%
\label{UoP_GT}}
\hfil
\subfloat[]{\includegraphics[width=1.1in]{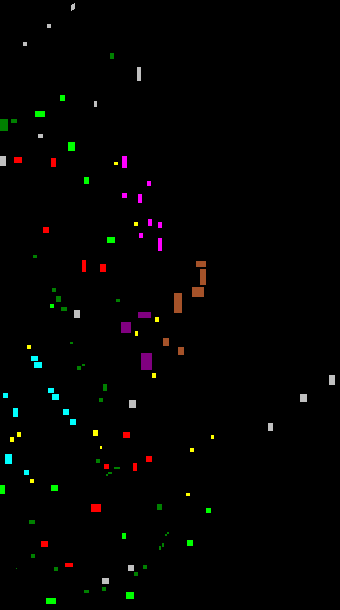}%
\label{UoP_Train}}
\hfil
\subfloat[]{\includegraphics[width=1.1in]{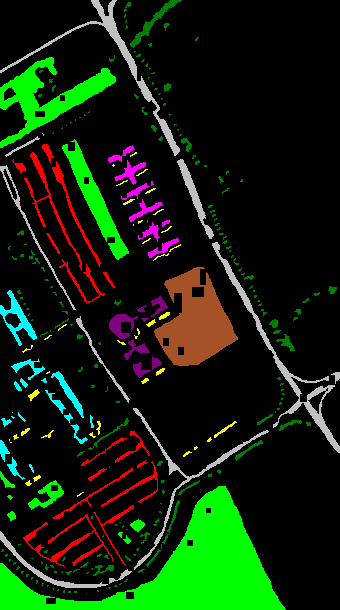}%
\label{UoP_Test}}
\hfil
\subfloat[]{\includegraphics[width=1.1in]{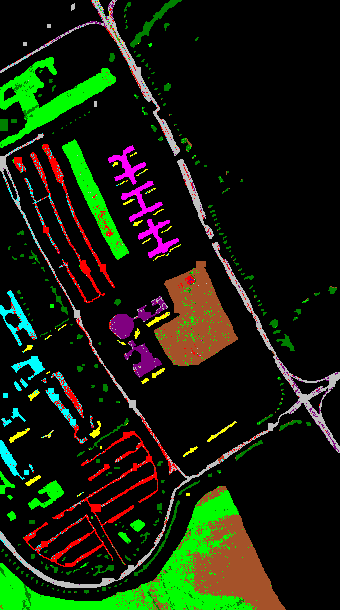}%
\label{UoP_SVM}}
\\
\subfloat[]{\includegraphics[width=1.1in]{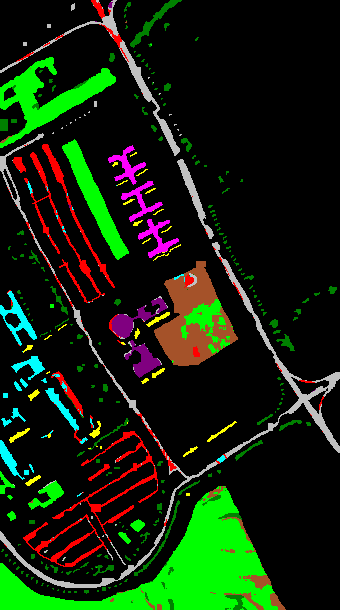}%
\label{UoP_CSVM}}
\hfil
\subfloat[]{\includegraphics[width=1.1in]{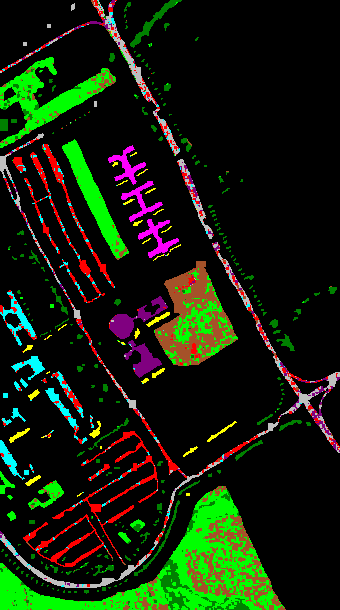}%
\label{UoP_SOMP}}
\hfil
\subfloat[]{\includegraphics[width=1.1in]{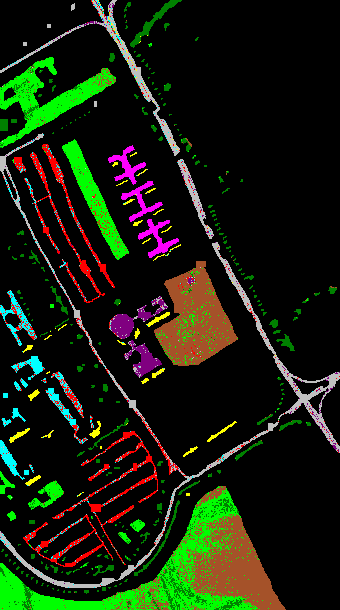}%
\label{UoP_SDL}}
\hfil
\subfloat[]{\includegraphics[width=1.1in]{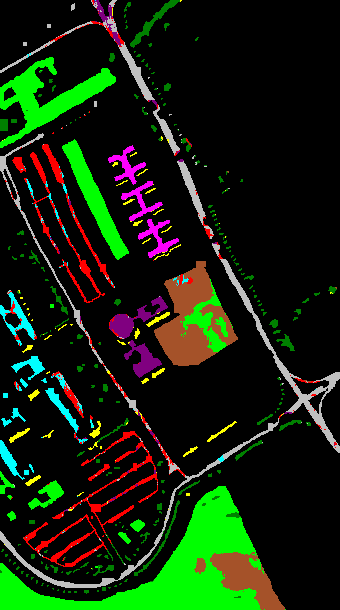}%
\label{UoP_CDL}}
\hfil
\subfloat[]{\includegraphics[width=1.1in]{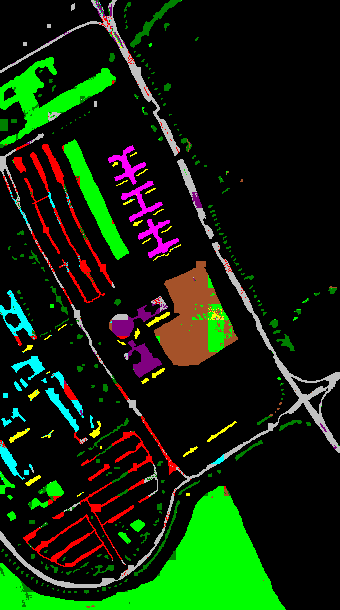}%
\label{UoP_SCDL}}
\caption{University of Pavia image. (a) Three-band color composite (bands 50,27, and 17). (b) Ground-truth. (c) Training data. (d) Test data. Classification maps obtained by (e) SVM, (f) CSVM$_{\mu,\sigma}$, (g) SOMP, (h) SDL, (i) CDL$_{\mu,\sigma}$, and (j) SCDL colored according to Table \ref{UoPCoP_Classes}.}
\label{UoP_image}
\end{figure*}

\begin{figure}[!t]
\centering
\subfloat[]{\includegraphics[width=1.1in]{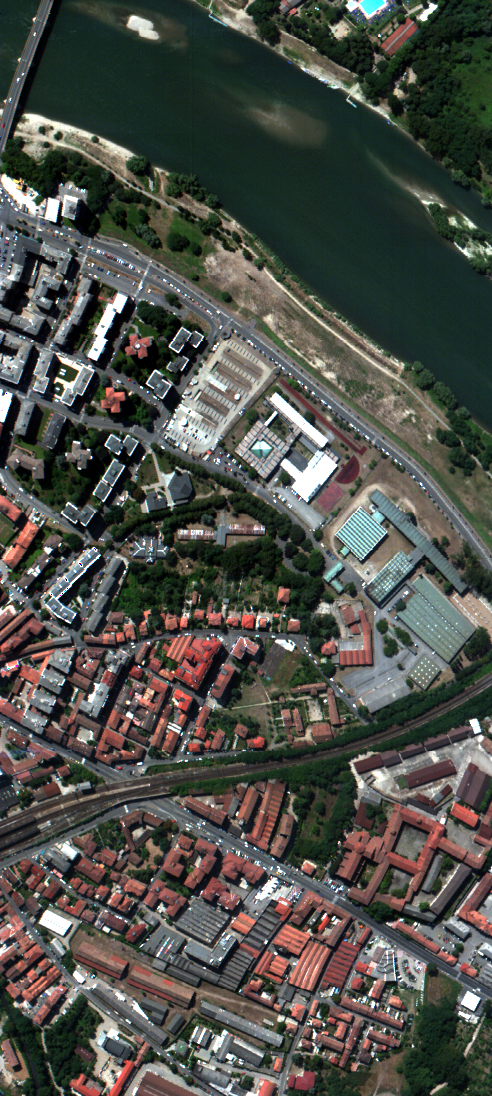}%
\label{PC_orig}}
\hfil
\subfloat[]{\includegraphics[width=1.1in]{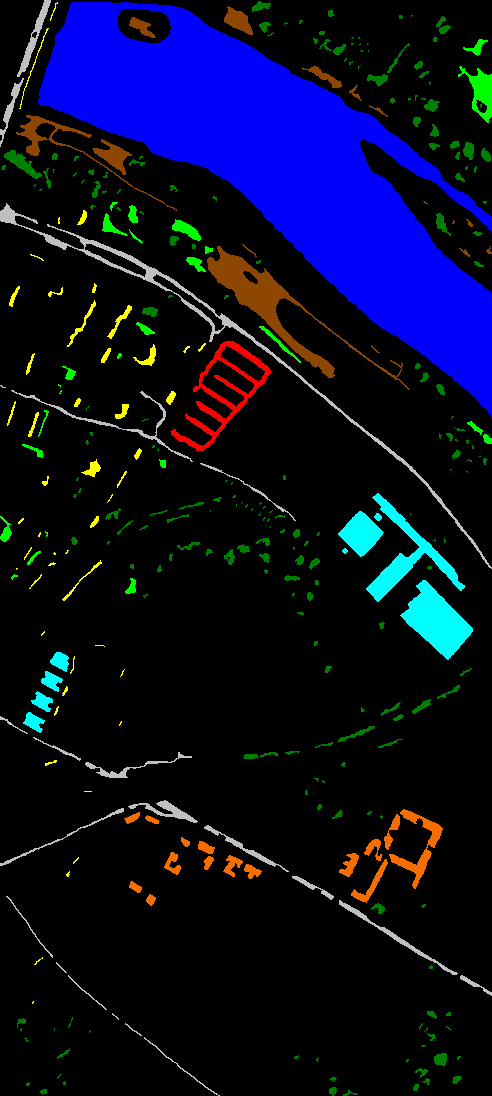}%
\label{PC_GT}}
\hfil
\subfloat[]{\includegraphics[width=1.1in]{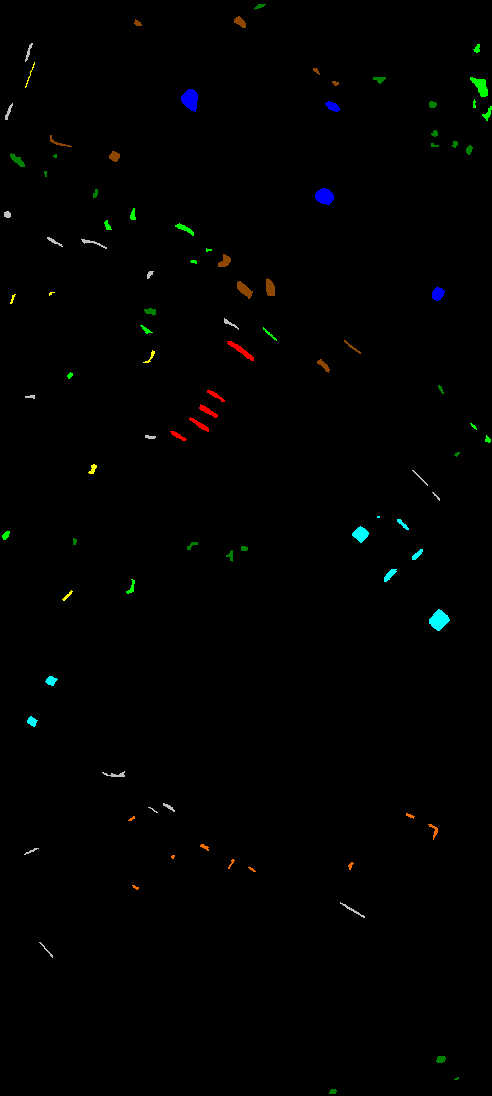}%
\label{PC_Train}}
\caption{Pavia Center image. (a) Three-band color composite (bands 50,27, and 17). (b) Ground-truth (c)  Train data colored according to Table \ref{UoPCoP_Classes}.}
\label{PC_image}
\end{figure}

ROSIS urban data refers to two datasets, namely University of Pavia (Fig. \ref{UoP_image}), and Center of Pavia (Fig. \ref{PC_image}), which were collected in 2003 by the ROSIS sensor with a spatial resolution of 1.3 m/pixel in 115 spectral bands covering 0.43 to 0.86 $\mu$m \cite{Plaza09}. For both images, specific test and train sets are used for classification as in \cite{Tarabalka09,Plaza09,Chen11}. Fig. \ref{UoP_Train} and Fig. \ref{PC_Train} show the training data for these images. 

Table \ref{UoPCoP_Classes} shows the 9 ground-truth classes and test and train sets of the University of Pavia image, which consists of $610\times340$ pixels in 103 spectral bands after 12 noisy bands were removed. Around 9\% of the data are used for training, leaving 91\% for testing. Experimental results for this image are reported in Table \ref{UoP_Accuracy} with classification maps for most classifiers depicted in Fig. \ref{UoP_image}. Classification accuracies for this image are notably lower than the Indian Pines or Pavia Center images as is confirmed by \cite{Plaza09,Chen11}. As noted earlier, SVM and SDL provide poor results since they only use the spectral characteristics. Unremarkably, CSVM and CKSVM provide the exact same results because the kernel weighting coefficient favors the contextual kernel in the cross-validation process of CKSVM. SOMP attains poor results because the training set is made up of small patches (Fig. \ref{UoP_Train}) and thus most likely the window surrounding a pixel contains no training samples. SCDL provides the best overall accuracy due to its ability to capture both spectral and contextual information. For further evaluation we chose random train and test data from the dataset and compared the overall classification accuracy of CSVM$_\mu$, SOMP, CDL$_\mu$, and SCDL. The results are found in Table \ref{UoP_Train_Accuracy} where the average accuracy and standard deviation is reported for 10 runs. Since for this experiment the training data is uniformly sampled from the entire image (unlike Fig. \ref{UoP_Train}) the obtained accuracies are higher than those reported in Table \ref{UoP_Accuracy}. Again, the results show that SCDL performs better for urban scenes, even when the training data is as low as 1\%.

\begin{table}[!t]
\renewcommand{\arraystretch}{1}
 %if using array.sty, it might be a good idea to tweak the value of
 %\extrarowheight %as needed to properly center the text within the cells
\caption{University of Pavia and Center of Pavia Ground-Truth Classes and Train/Test Sets }
\label{UoPCoP_Classes}
\centering
% Some packages, such as MDW tools, offer better commands for making tables
% than the plain LaTeX2e tabular which is used here.
\includegraphics{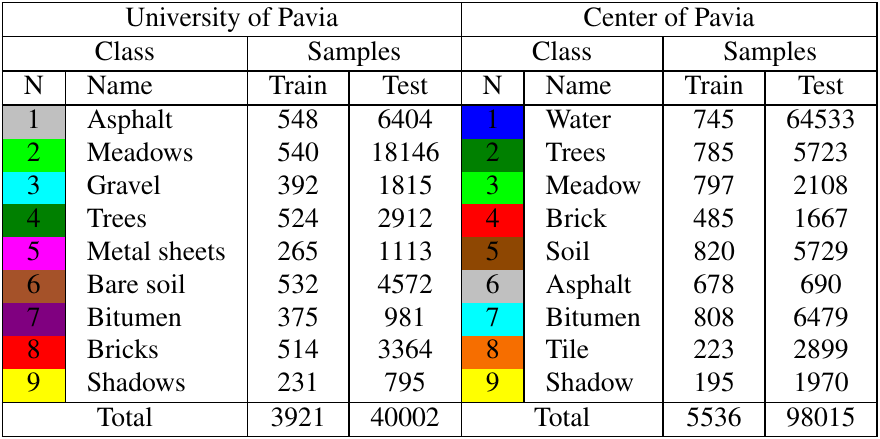}
%\begin{tabular}{|c|l|c|c|c|l|c|c|}
%\hline
%\multicolumn{4}{|c|}{University of Pavia} & \multicolumn{4}{c|}{Center of Pavia} \\
%\hline
%\multicolumn{2}{|c|}{Class} & \multicolumn{2}{c|}{Samples} & \multicolumn{2}{c|}{Class} & \multicolumn{2}{c|}{Samples} \\
%\hline
% N & Name & Train & Test & N & Name & Train & Test\\
%\hline
%1 \cellcolor{UoP1}& Asphalt & 548 & 6404 & 1 \cellcolor{PC1}& Water & 745 & 64533\\
%2 \cellcolor{UoP2}& Meadows & 540 & 18146 & 2 \cellcolor{PC2}& Trees & 785 & 5723\\
%3 \cellcolor{UoP3}& Gravel & 392 & 1815 & 3 \cellcolor{PC3}& Meadow & 797 & 2108\\
%4 \cellcolor{UoP4}& Trees & 524 & 2912 & 4 \cellcolor{PC4}& Brick & 485 & 1667\\
%5 \cellcolor{UoP5}& Metal sheets & 265 & 1113 & 5 \cellcolor{PC5}& Soil & 820 & 5729\\
%6 \cellcolor{UoP6}& Bare soil & 532 & 4572  & 6 \cellcolor{PC6}& Asphalt & 678 & 690\\
%7 \cellcolor{UoP7}& Bitumen & 375 & 981 & 7 \cellcolor{PC7}& Bitumen & 808 & 6479\\
%8 \cellcolor{UoP8}& Bricks & 514 & 3364 & 8 \cellcolor{PC8}& Tile & 223 & 2899\\
%9 \cellcolor{UoP9}& Shadows & 231 & 795 & 9 \cellcolor{PC9}& Shadow & 195 & 1970\\
%\hline
%\multicolumn{2}{|c|}{Total} & 3921 & 40002 & \multicolumn{2}{c|}{Total} & 5536 & 98015\\
%\hline\end{tabular}
\end{table}

The Pavia Center image consists of $1096\times492$ pixels in 102 spectral bands, after 13 noisy bands are removed. Table \ref{UoPCoP_Classes} shows the 9 ground-truth classes and test and train sets of the Pavia Center image. The classification results may be seen in Table \ref{PaviaC_Accuracy}. Similar to the previous image, SCDL obtains the best results, and CSVM and CKSVM perform alike. The previous two images also show the significance of second order moments for HSI classification in urban areas where the classes are usually scattered small regions in the image.

%\begin{table}[!t]
%% increase table row spacing, adjust to taste
%\renewcommand{\arraystretch}{1}
% %if using array.sty, it might be a good idea to tweak the value of
% %\extrarowheight %as needed to properly center the text within the cells
%\caption{Pavia Center Ground-Truth Classes and Train/Test Sets }
%\label{PaviaC_Classes}
%\centering
%% Some packages, such as MDW tools, offer better commands for making tables
%% than the plain LaTeX2e tabular which is used here.
%\begin{tabular}{|c|l||c|c|}
%\hline
%\multicolumn{2}{|c||}{Class} & \multicolumn{2}{c|}{Samples} \\
%\hline
%No & Name & Train & Test\\
%\hline
%1 \cellcolor{PC1}& Water & 745 & 64533 \\
%2 \cellcolor{PC2}& Trees & 785 & 5723 \\
%3 \cellcolor{PC3}& Meadow & 797 & 2108 \\
%4 \cellcolor{PC4}& Brick & 485 & 1667 \\
%5 \cellcolor{PC5}& Soil & 820 & 5729 \\
%6 \cellcolor{PC6}& Asphalt & 678 & 6907 \\
%7 \cellcolor{PC7}& Bitumen & 808 & 6479 \\
%8 \cellcolor{PC8}& Tile & 223 & 2899 \\
%9 \cellcolor{PC9}& Shadow & 195 & 1970 \\
%\hline
%Total & & 5536 & 98015\\
%\hline\end{tabular}
%\end{table}

\begin{table*}[!t]
\renewcommand{\arraystretch}{1}
 %if using array.sty, it might be a good idea to tweak the value of
 %\extrarowheight %as needed to properly center the text within the cells
\caption{Classification Accuracy (\%) on the University of Pavia Test Set for Different Classifiers}
\label{UoP_Accuracy}
\centering
% Some packages, such as MDW tools, offer better commands for making tables
% than the plain LaTeX2e tabular which is used here.
\includegraphics{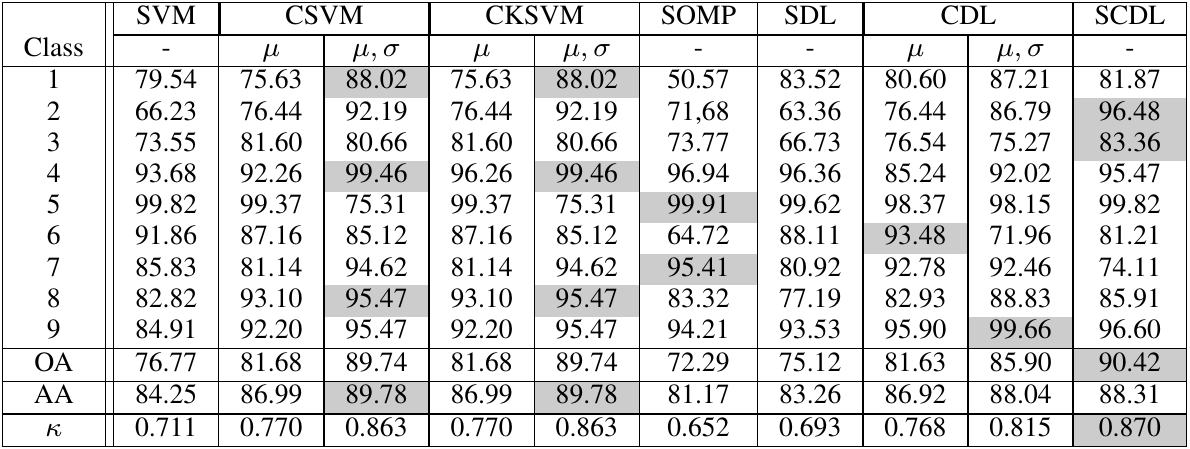}
\end{table*}

\begin{table}[!t]
\renewcommand{\arraystretch}{1}
 %if using array.sty, it might be a good idea to tweak the value of
 %\extrarowheight %as needed to properly center the text within the cells
\caption{Classification Accuracy (\%) on the University of Pavia dataset with different number of training data}
\label{UoP_Train_Accuracy}
\centering
% Some packages, such as MDW tools, offer better commands for making tables
% than the plain LaTeX2e tabular which is used here.
\includegraphics[width=3.45in]{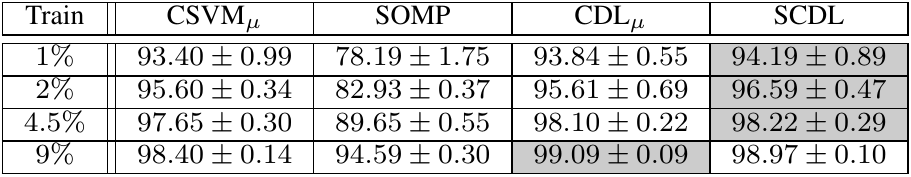}
\end{table}

\begin{table*}[!t]
\renewcommand{\arraystretch}{1}
 %if using array.sty, it might be a good idea to tweak the value of
 %\extrarowheight %as needed to properly center the text within the cells
\caption{Classification Accuracy (\%) on the Pavia Center Test Set for Different Classifiers}
\label{PaviaC_Accuracy}
\centering
% Some packages, such as MDW tools, offer better commands for making tables
% than the plain LaTeX2e tabular which is used here.
\includegraphics{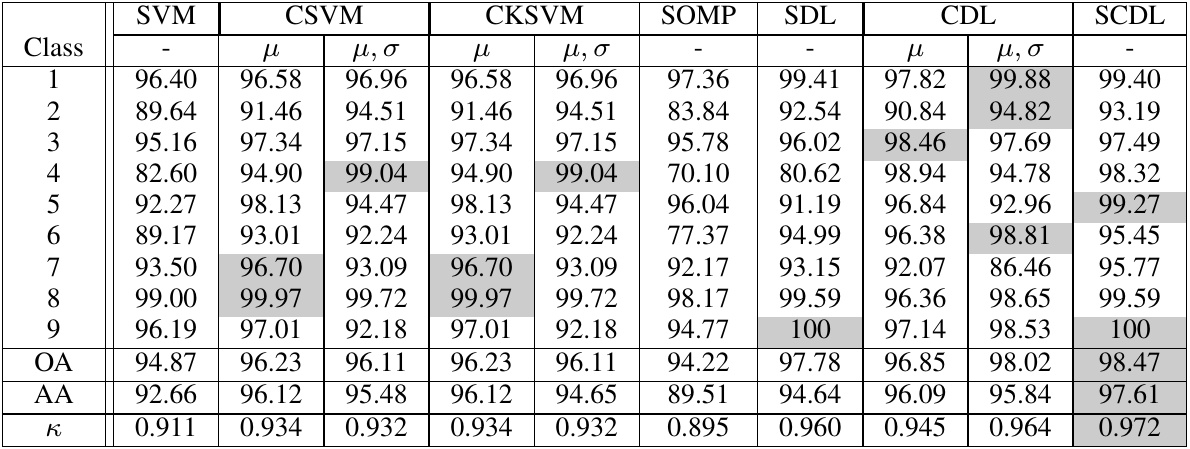}
\end{table*}

\section{Conclusion}
In this paper, we have investigated dictionary learning algorithms based on models of hyperspectral data for HSI classification. The fundamental idea of the models is to represent a hyperspectral sample with a linear combination of a few basic elements learned from the data. The spectral samples are classified using a linear SVM trained on the coefficients of this linear combination, known as the sparse representation. The models were also analyzed from a probabilistic viewpoint. The algorithms were used to exploit either one or both spectral and contextual information for HSI classification. Experiments on real HSI data confirmed the effectiveness of the models for HSI classification.

Many directions of future research are possible. The simple linear SVM yields accurate results, but takes little advantage of the sparsity of the representations. Also, we have observed that the sparse representations of CDL and SCDL are linear separable. Testing the models in conjunction with other classifiers might help us better exploit this potential. There exists a line of supervised dictionary learning algorithms \cite{Zhang10,Mairal12} which take advantage of the labels of the training instances to learn the sparse representations in a discriminative manner. To pursue this path we will need to deal with their expensive computation cost. We are also interested in testing the models for semisupervised classification of hyperspectral images, for which transductive SVMs \cite{Bruzzone06} seem to be a plausible choice. Regarding issues other than classification, it would be appealing to define the contextual groups using a smarter algorithm that takes advantage of both spectral and contextual characteristics of HSI.

% or


%\appendices
%% you can choose not to have a title for an appendix
%% if you want by leaving the argument blank
%\section{}

\section*{Acknowledgment}
The authors would like to thank the University of Pavia and Prof. Paolo Gamba for kindly providing the ROSIS images of University of Pavia and Center of Pavia and Prof. Landgrebe for the AVIRIS data. Finally, we would also like to thank Dr. Camps-Valls of the University of Valencia, Spain, for helpful discussions regarding the composite kernel SVM.

\ifCLASSOPTIONcaptionsoff
  \newpage
\fi



\bibliographystyle{IEEEtran}
\bibliography{IEEEabrv,./DL4HIC}

%\begin{thebibliography}{1}
%\end{thebibliography}

%\begin{biography}[{\includegraphics[width=1in,height=1.25in,clip,keepaspectratio]{mshell}}]{Michael Shell}
\vspace*{-2\baselineskip}

\begin{IEEEbiography}[{\includegraphics[width=1in,height=1.25in,clip,keepaspectratio]{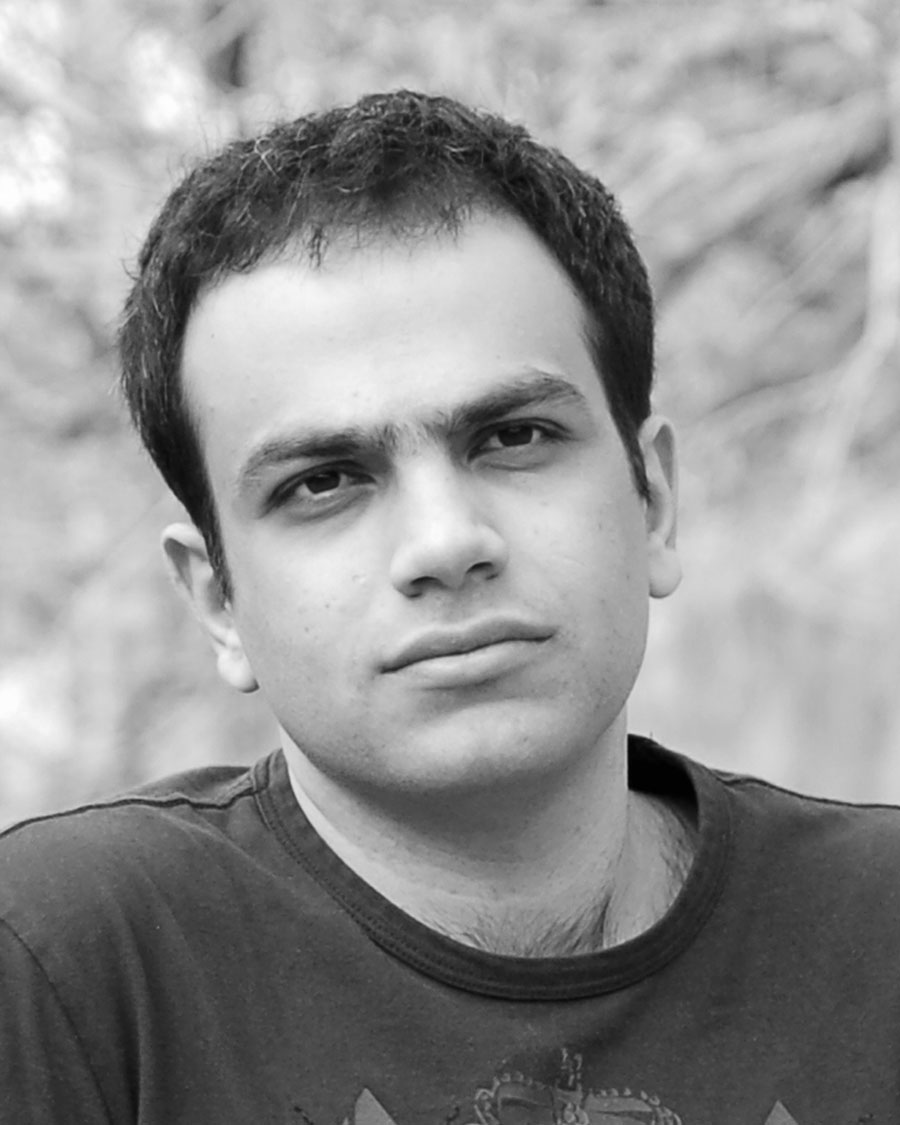}}]{Ali Soltani-Farani}
received his M.Sc. in Computer Architecture from Sharif University of Technology, Tehran, Iran, in 2010 and a B.Sc. in Hardware Engineering from the same university in 2008. He is currently working toward the Ph.D. degree in the Department of Computer Engineering at Sharif University of Technology.

%In 2009 he worked as a Researcher at Asr Guyesh Pardaz. 
From 2012 he works as a Technical Consultant at the Value Added Services Laboratory (VASL) and is an adjunct lecturer at Sharif University of Technology, Tehran, Iran. His current research interests include structured sparse representations, dictionary learning, and their application to signal and image processing.
\end{IEEEbiography}
\vspace*{-2\baselineskip}
%\newpage
\begin{IEEEbiography}[{\includegraphics[width=1in,height=1.25in,clip,keepaspectratio]{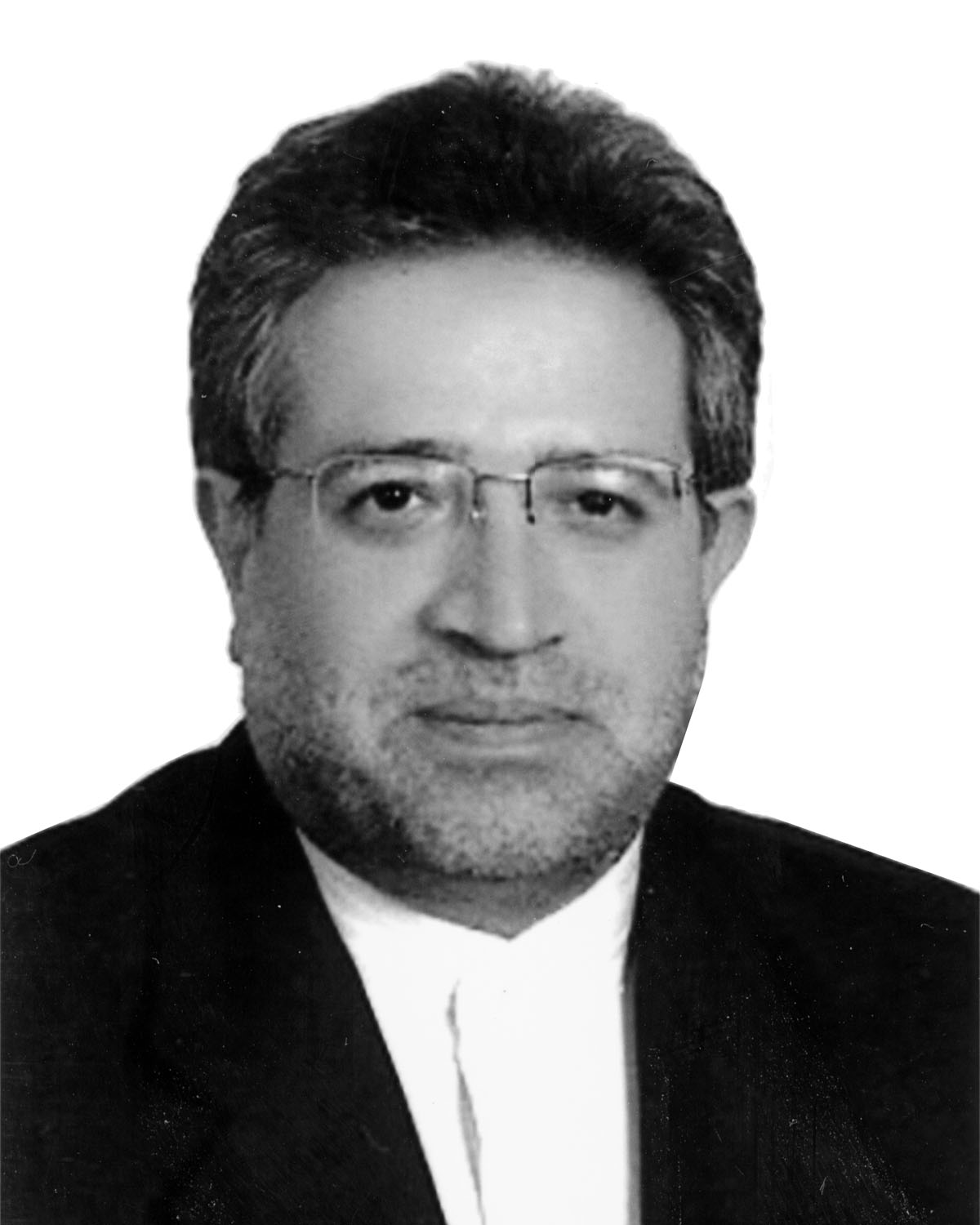}}]{Hamid R. Rabiee}
(SM'07) received his B.S. (1987) and M.S. (1989) degrees (with great distinction) in Electrical Engineering from CSULB, USA, his EEE in Electrical and Computer Engineering from USC, USA, and his Ph.D. (1996) in Electrical and Computer Engineering from Purdue University, West Lafayette, USA.

From 1993 to 1996 he was a Member of Technical Staff at AT\&T Bell Laboratories. From 1996 to 1999 he worked as a Senior Software Engineer at Intel Corporation. He was also with PSU, OGI, and OSU Universities as an adjunct professor of Electrical and Computer Engineering from 1996 to 2000. Since September 2000, he has joined Sharif University of Technology (SUT), Tehran, Iran. He is the founder of Sharif University Advanced Information and Communication Technology Research Center (AICT), Advanced Technologies Incubator (SATI), Digital Media Laboratory (DML) and Mobile Value Added Services Laboratory (VASL). He is currently a Professor of Computer Engineering at Sharif University of Technology and the Director of AICT, DML and VASL. He has been the initiator and director of national and international level projects in the context of UNDP International Open Source Network (IOSN) and Iran National ICT Development Plan.

Prof. Rabiee has received numerous awards and honors for his industrial, scientific and academic contributions. He has acted as chairman in a number of national and international conferences, and holds three patents.
\end{IEEEbiography}

%\newpage
\vspace*{-2\baselineskip}

\begin{IEEEbiography}[{\includegraphics[width=1in,height=1.25in,clip,keepaspectratio]{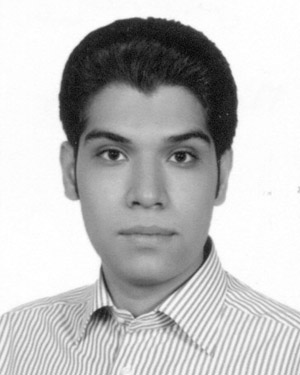}}]{Seyyed Abbas Hosseini}
received his B.Sc. in Software Engineering from Sharif University of Technology, Tehran, Iran, in 2012 and is currently working towards his M.Sc. degree in the Department of Computer Engineering at Sharif University of Technology.

%In 2011 he spent a summer internship at the Probabilistic Machine Learning Laboratory, Echole Polytechnique Federale De Lausanne, Lausanne, Switzerland.
His current research interests include the application of dictionary learning and metric learning to image classification.
\end{IEEEbiography}

%\begin{IEEEbiography}[{\includegraphics[width=1in,height=1.25in,clip,keepaspectratio]{jalali.jpg}}]{Atefeh Jalali}
%received her M.Sc. in Artificial Intelligence from Tabriz University, Tabriz, Iran, in 2012 and a B.Sc. in Software Engineering from Sajjad Institute for Higher Education, Mashhad, Iran, in 2009.
%
%From 2011 to 2013 she worked as a Research Assistant at the Digital Media Laboratory (DML) at Sharif University of Technology, Tehran, Iran. Her current research interests include sparse representations, machine learning, and their application to signal and image processing.
%\end{IEEEbiography}


%\vfill


\end{document}